# Process-Aware AI for Rainfall–Runoff Modeling: A Mass-Conserving Neural Framework with Hydrological Process Constraints

Mohammad A. Farmani[1], Hoshin V. Gupta[1], Ali Behrangi[1,2], Muhammad Jawad[1], Sadaf Moghisi[1], Guo-Yue Niu[1]

[1]Department of Hydrology and Atmospheric Sciences, University of Arizona, Tucson, AZ, USA,
[2]Department of Geosciences, University of Arizona, Tucson, AZ, USA,

Contact: farmani@arizona.edu
Submitted to arXiv

**Key Points:**

- Embedding hydrological process constraints within a mass-conserving AI framework improves interpretable rainfall–runoff prediction.
- Incorporating vertical drainage strongly improves performance in arid and snow-dominated basins but reduces skill in rainfall-dominated regions.
- Process-aware AI models can approach deep-learning predictive skill while retaining physically-interpretable storage–flux dynamics.

**Abstract**

Machine learning models can achieve high predictive accuracy in hydrological applications but often lack physical interpretability. The *Mass-Conserving Perceptron* (MCP) provides a physics-aware artificial intelligence (AI) framework that enforces conservation principles while allowing hydrological process relationships to be learned from data.

In this study, we investigate how progressively embedding physically meaningful representations of hydrological processes within a single MCP storage unit improves predictive skill and interpretability in rainfall–runoff modeling. Starting from a minimal MCP formulation, we sequentially introduce bounded soil storage, state-dependent conductivity, variable porosity, infiltration capacity, surface ponding, vertical drainage, and nonlinear water-table dynamics.

The resulting hierarchy of process-aware MCP models is evaluated across 15 catchments spanning five hydroclimatic regions of the continental United States using daily streamflow prediction as the target. Results show that progressively augmenting the internal physical structure of the MCP unit generally improves predictive performance. The influence of boundary conditions is strongly hydroclimate dependent: vertical drainage substantially improves model skill in arid and snow-dominated basins but reduces performance in rainfall-dominated regions, while surface ponding has comparatively small effects.

The best-performing MCP configurations approach the predictive skill of a Long Short-Term Memory benchmark while maintaining explicit physical interpretability. These results demonstrate that embedding hydrological process constraints within AI architectures provides a promising pathway toward interpretable and process-aware rainfall–runoff modeling.

**Plain Language Summary**

Machine learning (ML) models are increasingly used to predict river flow because they can learn complex patterns from data. However, these models often behave like "*black boxes*," making it difficult to understand their representation of how water moves through a watershed. In contrast, traditional physical-based (PB) hydrological models are easier to interpret (because they represent physical processes such as soil storage and runoff generation) but may be less capable of expressing complex relationships.

In this study, we combine ideas from both ML and PB approaches by developing a hybrid model that conserves water mass and includes physically meaningful hydrological processes, and progressively add processes such as limits on soil water storage, infiltration controls, surface ponding, and deep drainage. Testing on 15 river basins that represent a wide range of climate conditions across the United States reveals that adding key physical processes improves streamflow predictions, with process importance varying by region. Representing deep drainage substantially improves predictions in arid and snow-dominated basins but can reduce performance in humid regions. These findings show that AI models can remain physically-interpretable while achieving predictive skill comparable to modern deep-learning approaches.

# 1 Introduction

[1] Rainfall–runoff (RR) models provide the foundation for understanding and predicting catchment-scale hydrological responses, while supporting applications such as flood forecasting, drought assessment, and water resources management. Over the past several decades, hydrological modeling has evolved from spatially-lumped conceptual approaches toward semi-distributed and physically-based frameworks that explicitly represent processes such as infiltration, evapotranspiration, and subsurface flow (*Gupta et al., 1998; Burnash, 1973; Bergström, 1995*). More recent high-resolution distributed and land surface models incorporate spatial heterogeneity and coupled water–energy dynamics, extending applicability across diverse hydroclimatic regimes *(Burnash, 1973; Niu et al., 2011; Agnihotri et al., 2023; Solomon et al., 2024; Farmani et al., 2024; Farmani et al., 2025*).

[2] Despite these advances, RR modeling continues to face a fundamental challenge: achieving predictive accuracy while maintaining physical realism and interpretability. Physical-conceptual models rely on simplified process representations and parameter calibration, which can limit transferability under changing climatic conditions and across catchments (*Beven, 2006; Gupta et al., 2009*). Physically-based models offer greater process realism but require extensive data, substantial computational resources, and their parameterizations remain uncertain (*Clark et al., 2015a; Clark et al., 2015b; Addor and Melsen, 2019; Nearing et al., 2021*). Model performance is further influenced by input uncertainty, parameter equifinality, and incomplete representation of key processes such as subsurface flow and snow dynamics (*Beven, 2006; Kirchner, 2006; Moghisi et al., 2024*). Consequently, hydrological modeling remains characterized by trade-offs among complexity, interpretability, and predictive skill.

[3] To address limitations of traditional approaches, data-driven methods — based particularly in machine learning (ML) — have gained increasing attention. Deep learning (DL) architectures, including recurrent neural networks and *Long Short-Term Memory* (LSTM) models, learn nonlinear relationships between meteorological inputs and streamflow without explicitly representing individual process equations. These models often achieve strong predictive performance across diverse basins and large-sample datasets such as CAMELS (*Kratzert et al., 2018; Kratzert et al., 2019*), and their ability to capture temporal dependencies and leverage large training datasets has demonstrated the potential of ML to advance regional and continental-scale hydrologic prediction.

[4] However, data-driven models can function as "*black boxes*", with their internal representations being difficult to interpret in physical terms, raising concerns about extrapolation under non-stationary climate conditions and limiting their usefulness for scientific understanding (*Shen, 2018; Reichstein et al., 2019; Beven, 2020*). ML models may also violate physical constraints, such as conservation of mass, or compensate for structural deficiencies through parameter adjustments that are uninterpretable (*Karpatne et al., 2017; Willard et al., 2022*). These limitations highlight the need for modeling frameworks that achieve the predictive capability of ML while maintaining physical consistency and interpretability.

[5] *Physics-Informed Machine Learning* (PIML) represents one response to this need. PIML approaches incorporate physical laws, constraints, or structural knowledge into the learning process to guide model behavior (*Karpatne et al., 2017; Reichstein et al., 2019; Willard et al., 2022*). For example, *Physics-Informed Neural Networks* (PINNs) embed governing equations within the loss function to enforce conservation principles (*Raissi et al., 2019*). In hydrology, related efforts enforce mass balance, constrain process behavior, or integrate neural networks with process-based components (*Read et al., 2019; Feng et al., 2020; Jia et al., 2021*). While these approaches improve physical consistency, physical knowledge is often introduced as soft constraints or external regularization, rather than being embedded directly into the computational structure. As a result, learned internal states and fluxes do not necessarily correspond to physically meaningful quantities.

[6] *Differentiable Hydrology* extends this line of development by reformulating process-based models as mathematically-differentiable systems that can be trained using gradient-based optimization (*Tsai et al., 2021; Shen et al., 2023*). This framework enables end-to-end parameter learning and hybrid architectures that combine mechanistic equations with neural network components. Dynamic parameterization has been proposed to represent time-varying processes such as vegetation dynamics and snow interactions (*Feng et al., 2022; Shen et al., 2023; Song et al., 2025*). Although these methods improve predictive performance, dynamically learned parameters may lack physical identifiability or compensate for structural deficiencies, limiting interpretability and mechanistic insight.

[7] The *Mass-Conserving Perceptron* (MCP) provides an alternative approach that embeds physical constraints directly within the computational architecture. It was introduced to address a key limitation of DL in hydrology: while gated recurrent neural networks achieve high predictive accuracy, their internal states and fluxes lack physical interpretability. *Wang and Gupta (2024a)* reformulated gated recurrent units into mass-conserving computational units whose state variables and fluxes explicitly satisfy conservation principles. MCP units are isomorphically related to gated RNN cells but are recast so that storage and flux components correspond to physically-interpretable quantities, enabling hypothesis testing within a differentiable framework.

[8] Subsequent work extended this formulation by assembling MCP units into directed-graph architectures resembling conceptual RR models (*Wang and Gupta, 2024b*). These studies demonstrated that mass-conserving architectures can achieve competitive predictive performance while retaining interpretability and enabling systematic exploration of structural hypotheses. More recent developments emphasized parsimony and minimally-optimal representations, arguing that model complexity should be aligned with dominant hydrological processes (*Wang and Gupta, 2025*). A large-sample CONUS-wide evaluation further showed that MCP-based interpretable models can approach LSTM-level predictive skill while highlighting the trade-off between accuracy and architectural complexity across hydroclimatic regimes (*Wang and Gupta, 2025*).

[9] While the MCP framework provides an appealing foundation for interpretable hydrological ML, an important question remains unresolved:

*How much physical process representation should be embedded within the model structure?*

A minimal MCP formulation can represent storage–flux dynamics in a highly flexible way, but it may lack explicit representations of key hydrological processes such as infiltration capacity limits, ponding, groundwater drainage, and soil-water storage constraints. Conversely, introducing too many process representations may increase structural complexity without improving predictive skill.

[10] In this study, we investigate how progressively embedding physically-meaningful hydrological processes within an MCP architecture influences RR modeling performance. Starting from a minimal MCP formulation, we sequentially introduce additional physical constraints and process representations to construct a *hierarchy of process-aware mass-conserving AI models*. These progressively-augmented MCP architectures incorporate hydrological mechanisms including bounded soil storage, state-dependent hydraulic conductivity, infiltration capacity limits, surface ponding, vertical drainage, and nonlinear water-table behavior. The resulting models represent a continuum between flexible data-driven learning and physically-interpretable hydrological modeling.

[11] Performance of the models is evaluated across 15 catchments spanning five major hydroclimatic regimes of the continental United States (CONUS), including arid, snow-dominated, and rainfall-dominated systems. Predictive performance is assessed using daily streamflow simulation and compared against a benchmark LSTM model trained with the same meteorological inputs. In addition to evaluating overall predictive skill, we examine how model behavior varies across flow regimes, how boundary-

condition assumptions influence performance, and how the learned storage–flux relationships reflect hydrological processes.

[12] The central hypothesis of this work is that:

*Embedding physically-meaningful hydrological processes within a mass-conserving AI architecture can improve predictive skill while preserving interpretability.*

By systematically analyzing the influence of progressive process augmentation across diverse hydroclimatic regimes, this study aims to clarify how process-aware ML frameworks can bridge the gap between traditional hydrological models and purely data-driven approaches.

# 2 Methods

[13] This study develops a physically-interpretable hydrological modeling framework based on the MCP architecture and evaluates how progressively embedding hydrologically-meaningful processes within a single storage unit influences model behavior and predictive skill. The framework is constructed by systematically augmenting the internal dynamics of an MCP storage unit to represent key hydrological mechanisms, resulting in a hierarchy of progressively more physically-informed models. In addition to internal process representations, alternative boundary-condition configurations are introduced to account for *surface ponding* and *vertical drainage*. Together, these structural components define a structured experimental space that enables systematic investigation of how internal soil processes and boundary conditions influence rainfall–runoff dynamics.

[14] The following sub-sections first introduce the MCP framework used in this study (Section 2.1). We then describe:

(i) Progressive augmentation of the MCP storage unit to represent key hydrological processes (Section 2.2)

(ii) Representation of surface ponding and vertical drainage processes (Section 2.3)

(iii) The resulting model configuration space (Section 2.4), and

(iv) The datasets, experimental design, training procedures, and evaluation metrics used to assess model performance (Section 2.5).

## 2.1 The Mass-Conserving Perceptron (MCP)

[15] The MCP is a mathematically-differentiable computational unit designed to integrate physical consistency with data-driven learning by explicitly enforcing conservation laws within its architecture (*Wang and Gupta, 2024a*). Unlike conventional neural network units, which operate as abstract nonlinear transformations, the MCP is formulated as a storage–flux system in which the internal state represents a physically-interpretable storage variable and all incoming and outgoing fluxes are explicitly accounted for.

[16] **Figure 1** illustrates the structure of a single MCP unit. The system is characterized by a storage state $X_t$ that evolves in discrete time in response to external inputs and outgoing fluxes. The temporal evolution of storage is governed by a mass balance equation:

$$X_{t+1} = X_t + U_t G_t^U - O_t - L_t \quad \text{(Eq 1)}$$

where $X_{t+1}$ denotes storage at the next time step, $U_t$ represents external input (e.g., precipitation), $G_t^U$ is an input gate controlling the fraction of input entering storage, $O_t$ denotes output flux (e.g., streamflow), and $L_t$ represents loss flux (e.g., evapotranspiration).

[17] Outgoing fluxes are modeled as proportional to the current storage:

$$O_t = G_t^O X_t \quad \text{(Eq 2)}$$

$$L_t = G_t^L X_t \quad \text{(Eq 3)}$$

Substituting into the mass balance gives:

$$X_{t+1} = X_t + U_t G_t^U - G_t^O X_t - G_t^L X_t \quad \text{(Eq 4)}$$

which can be rewritten as:

$$X_{t+1} = G_t^R X_t + U_t G_t^U \quad \text{(Eq 5)}$$

where $G_t^R = 1 - G_t^O - G_t^L$ represents the retention gate. The gating variables $G_t^U$, $G_t^O$, and $G_t^L$ are constrained to lie within the interval [0,1], ensuring physically-meaningful partitioning of mass. By construction, the sum of outgoing and retained fractions equals unity, guaranteeing that total mass is conserved at each time step (*Wang and Gupta, 2024a*).

[18] The gating functions are parameterized using differentiable transformations. For example:

$$G_t^O = \sigma(b_O + a_O \widetilde{X_t}) \quad \text{(Eq 6)}$$

where $a_O$ and $b_O$ are learnable parameters and $\widetilde{X_t}$ is a scaled storage state. This structure makes the MCP isomorphically related to gated recurrent neural networks, while ensuring that internal states and fluxes retain direct physical interpretation (*Wang and Gupta, 2024b*). Because conservation is embedded directly in the computational architecture rather than imposed through external constraints, the MCP provides a physically-interpretable recurrent unit that remains fully differentiable and trainable.

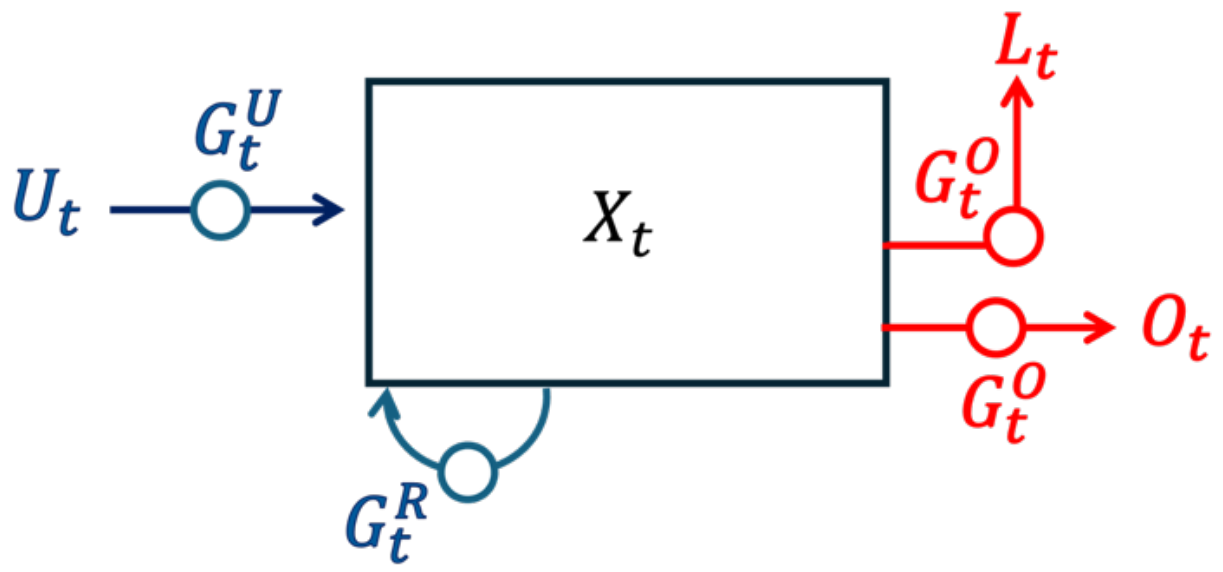

**Figure 1. Schematic of a single *Mass-Conserving Perceptron* (MCP) unit.** The internal state $X_t$ represents storage, which evolves according to a mass balance equation. Incoming input $U_t$ enters the storage through the input gate $G_t^U$. Outgoing fluxes are partitioned into output $O_t$ (controlled by $G_t^O$) and loss $L_t$ (controlled by $G_t^L$). The retention gate $G_t^R$ represents the fraction of storage retained between time steps. All fluxes are explicitly accounted for, ensuring conservation of mass at the unit level.

**Table 1. Key symbols used in the MCP framework and augmented model formulations.**

| Symbol | Description |
|---|---|
| $\boldsymbol{X_t}$ | MCP storage state |
| $\boldsymbol{U_t}$ | precipitation input |
| $\boldsymbol{O_t}$ | baseflow / discharge flux |
| $L_t$ | evapotranspiration loss |
| $R_t^{SE}$ | saturation-excess runoff |
| $R_t^{IE}$ | infiltration-excess runoff |
| $V_t$ | vertical drainage |
| $\theta_t$ | soil water storage |
| $\theta_{max}$ | maximum soil storage |
| $\theta_{min}$ | wilting storage |
| $\theta_a$ | available soil water |
| $H_{soil}$ | soil thickness |
| $\rho$ | porosity |
| $s_t$ | soil saturation state |
| $G_t^U$ | infiltration gate |
| $G_t^O$ | baseflow gate |
| $G_t^V$ | vertical drainage gate |

### 2.2 Progressive Physical Augmentation of the MCP Storage Unit

[19] In its original formulation, the MCP represents a generic mass-conserving storage unit that learns flux-partitioning directly from data. In this study, we progressively embed physically-meaningful hydrological processes within a single MCP storage unit in order to improve interpretability while preserving mathematical-differentiability and strict conservation of mass. The augmentation strategy is designed to isolate the influence of individual hydrological processes on model behavior. Each successive model architecture introduces a single additional process representation while preserving all previously introduced mechanisms. This design enables systematic evaluation of how increasing physical realism within a single MCP storage unit influences predictive skill and hydrological behavior. The resulting hierarchy consists of five progressively-augmented models (M1–M5) developed under a baseline configuration without surface ponding or vertical drainage (NP). Figure 2 illustrates the structural evolution of the MCP storage unit as additional hydrological processes are incorporated.

**Table2: Progressive process augmentation of the MCP storage unit under the baseline NP configuration.**

| Model | New process introduced | Hydrological interpretation |
|---|---|---|
| M1 | bounded soil bucket | conceptual soil bucket |
| M2 | State-dependent conductivity | moisture-controlled drainage |
| M3 | Variable porosity | volumetric soil storage capacity |
| M4 | Infiltration capacity & Horton runoff | infiltration-excess runoff generation |
| M5 | Nonlinear water table dynamics | partial saturation and internal redistribution |

### 2.2.1 *Baseline Soil Bucket Representation (M1)*

[20] We reinterpret the MCP storage state $X_t$ as soil water storage $\theta_t$, representing water held within an effective soil bucket. The soil bucket is defined by upper and lower storage bounds:

$$\theta_{max} = H_{soil} \cdot \rho, \quad r_p = \frac{\theta_{min}}{\theta_{max}} \qquad \text{(Eq 7)}$$

where $H_{soil}$ [L] is soil thickness, $\rho$ [–] is porosity, $\theta_{max}$ [L] is maximum storage capacity, and $\theta_{min}$ [L] is minimum storage corresponding to the wilting threshold. In this study, $H_{soil}$ is obtained from CAMELS static attributes, and $r_p$ is fixed at 0.1 across basins, so $\theta_{min} = 0.1\, \theta_{max}$.

[21] For M1 and M2, we set $\rho = 1$, yielding a conceptual soil "*tank*" representation. In later models we allow $\rho < 1$, which converts the tank into a volumetric soil-column representation by explicitly defining pore space.

[22] The degree of saturation is defined as:

$$s_t = \frac{\theta_t - \theta_{min}}{\theta_{max} - \theta_{min}} \qquad \text{(Eq 8)}$$

which normalizes soil water content to [0,1] and provides a physically-interpretable state variable used to drive gating functions.

[23] Accordingly, the M1 mass balance becomes:

$$\theta_{t+1} = \theta_t + U_t - O_t - L_t - R_t^{SE} \qquad \text{(Eq 9)}$$

where $U_t$ is precipitation input, $O_t$ is baseflow (slow drainage), $L_t$ is evapotranspiration loss, and $R_t^{SE}$ is saturation-excess runoff. Saturation-excess runoff occurs when incoming water pushes storage beyond $\theta_{max}$, with the excess routed to $R_t^{SE}$ to maintain physical bounds.

[24] Finally, the extractable water available for drainage and evapotranspiration is:

$$\theta_a = \theta_t - \theta_{min} \qquad \text{(Eq 10)}$$

### 2.2.2 *Representation of Horizontal-Drainage Baseflow in M1*

[25] In M1, baseflow is modeled using a constant horizontal-drainage conductivity parameter to isolate the effect of state-independent drainage rate:

$$O_t = G_t^O . \theta_a \text{ with } G_t^O = K_{sat} \qquad \text{(Eq 11)}$$

where $K_{sat}$ is a (learnable or prescribed) saturated hydraulic conductivity coefficient. This formulation produces drainage at a rate that is linearly proportional to available soil water.

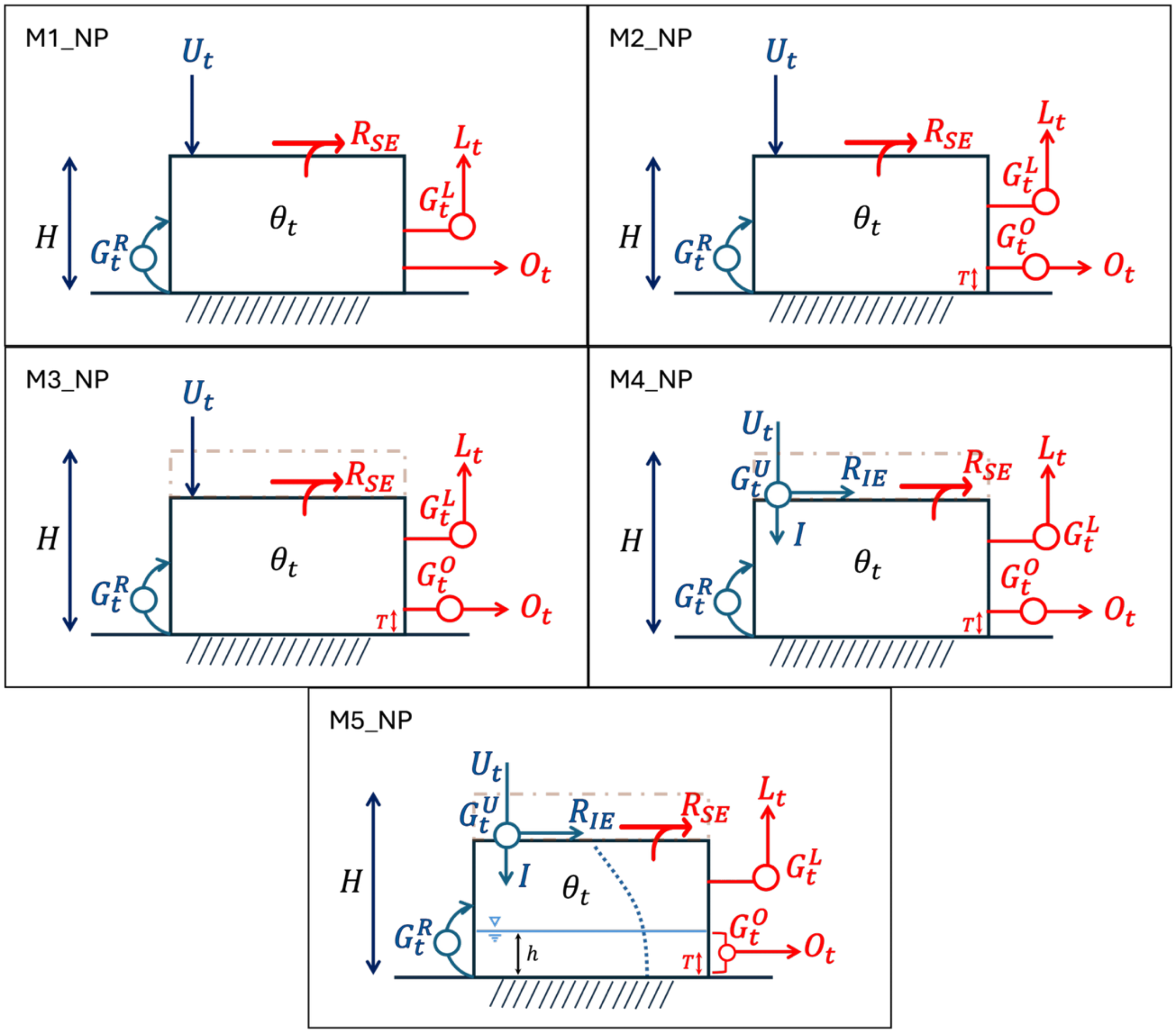

**Figure 2. Conceptual evolution of the MCP storage unit as progressively richer physical constraints are introduced (M1–M5) under the no-ponding, no-drainage (NP) configuration.** Starting from a bounded soil bucket (M1), subsequent models incorporate state-dependent conductivity (M2), volumetric soil capacity through porosity (M3), infiltration capacity and infiltration-excess runoff (M4), and nonlinear water-table dynamics representing partial saturation and internal redistribution (M5). The diagrams illustrate how physically-interpretable processes are embedded within the MCP storage unit while maintaining mass conservation.

### *2.2.3 Evapotranspiration as Root-Weighted Extraction*

[26] Evapotranspiration is represented consistently across all models and scenarios by converting atmospheric demand (PET) into soil water extraction limited by soil moisture availability and root distribution.

$$L_t = \min(\theta_a, E_{demand}) \quad \text{(Eq 12)}$$

$$E_{demand} = PET \,.\, R_t^{root} .\, (s_t)^{b_L} \quad \text{(Eq 13)}$$

where PET is potential evapotranspiration, $b_L$ is a moisture-stress exponent, and $R_t^{root}$ is a normalized root-uptake fraction.

[27] Although the experiments in this paper use a single soil layer, we formulate root uptake using a general N-layer form so the framework can be extended to multilayer representations, **Figure 3**. For layer $j$ with depth coordinate $z_j$, we define an un-normalized root fraction:

$$R_j^{frac} = \sigma(m_L . \breve{z}_j - c_L) \quad \text{(Eq 14)}$$

and normalize across $N$ layers:

$$R_t^{root} = \frac{R_j^{frac}}{\sum_{j=1}^{N} R_j^{frac})} \quad \text{(Eq 15)}$$

[28] **Figure 3** illustrates the normalized root profile for N=3.

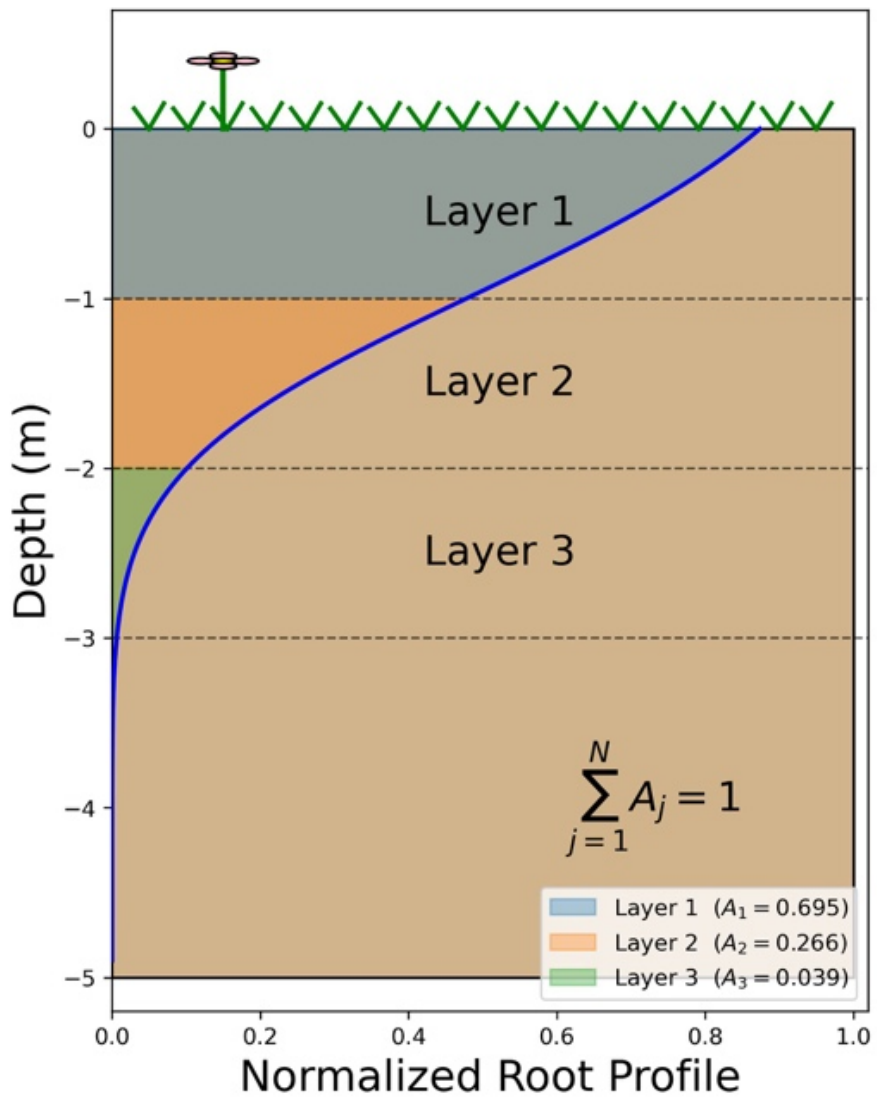

**Figure 3. Conceptual root distribution used to represent evapotranspiration extraction within the soil column.** The blue curve shows the normalized root density profile as a function of depth. The soil column is divided into three conceptual layers, each receiving a fraction $A_j$ of the total root uptake such that $\sum_{j=1}^{N} A_j = 1$. In this study the formulation is written for a general multilayer representation, although simulations are performed using a single effective layer.

### *2.2.4 State-Dependent Baseflow Via Moisture-Controlled Conductivity (M2–M5)*

[29] In real soils, hydraulic conductivity depends on moisture state through soil water pressure head. To represent this behavior without introducing unconstrained "*dynamic parameters*," we model conductivity as a *state-dependent gating function* that maps the (physically-interpretable) saturation state to an effective conductivity. For models M2–M5, baseflow is computed as:

$$O_t = G_t^O . D_t \quad \text{(Eq 16)}$$

$$G_t^O = K_{sat} . \sigma(a_O . \breve{x}_h - b_O) \quad \text{(Eq 17)}$$

where $a_O$ and $b_O$ are learnable gate parameters and $\breve{x}_h$ is a scaled water-table activation variable. Water-table height is defined as:

$$h = H * s_t^{b_w} \quad \text{(Eq 18)}$$

where $b_w$ is a water-table shape parameter. The active drainage zone depth is:

$$Dr = H - T \quad \text{(Eq 19)}$$

[30] The water-table activation variable becomes:

$$x_h = \max\left(0, \frac{h-T}{Dr}\right) \quad \text{(Eq 20)}$$

and drainable water is defined as:

$$D_t = \max(0, (h - T) * \frac{\theta_a}{H}) \quad \text{(Eq 21)}$$

where $T$ represents a threshold water-table height required to activate significant drainage. The use of $D_t$ ensures that baseflow depends on both (i) the extent of saturated storage above T and (ii) the available water $\theta_a$.

#### 2.2.5 *Water Table Representation (M5)*

[31] For M2–M4 we set $b_w = 1$, yielding a linear relationship between saturation and water-table height, consistent with common conceptualizations that assume uniform moisture distribution within a bucket. For M5 we allow $b_w > 1$, which produces a non-linear mapping and implies non-uniform vertical water distribution, with a partially saturated lower zone and an un-saturated upper zone (**Figure 2**, M5_NP). This modification enables a simple representation of a dynamic water table and internal redistribution within a single storage unit.

#### 2.2.6 *Variable Porosity (M3)*

[32] In M1 and M2, $\rho = 1$ results in a conceptual tank with water-holding capacity determined by $H_{soil}$. In M3 and higher, we allow $\rho < 1$, which explicitly represents *pore space* and converts the conceptual tank into a *volumetric soil layer*. Storage capacity therefore becomes directly dependent on soil porosity.

#### 2.2.7 *Infiltration Capacity and Infiltration-Excess Runoff (M4–M5)*

[33] To represent infiltration limitation and infiltration-excess runoff generation, precipitation is partitioned using an input gate:

$$I_t = U_t G_t^U \text{ where } G_t^U = \sigma(a_U \widetilde{s_t} - b_U) \quad \text{(Eq 22)}$$

$$R_t^{IE} = U_t G_t^{IE} \text{ where } G_t^{IE} = 1 - G_t^U \quad \text{(Eq 23)}$$

where $a_U$ and $b_U$ are learnable parameters and $\widetilde{s_t}$ is a scaled saturation state. This design enforces physically-interpretable behavior: as the soil approaches saturation, $G_t^U$ decreases, limiting infiltration and increasing infiltration-excess runoff. When the soil is dry, $G_t^U$ increases (maximum), allowing most incoming water to infiltrate, subject to available storage capacity (with any additional excess still contributing to $R_t^{SE}$ once $\theta_{max}$ is reached).

### 2.3 Representation of Surface Ponding (PND)

[34] To represent temporary surface water storage and delayed runoff generation, we extend the previous set of "*non-ponding*" models by introducing a surface ponding layer. This boundary-condition scenario is referred to as PND, and the resulting models are denoted M1_PND to M5_PND (**Figure 4**).

[35] A *surface storage state*, $S_t^{PND}$ is introduced to represent water temporarily stored above the soil column. The ponding storage evolves according to:

$$S_{t+1}^{PND} = S_t^{PND} + U_t - I_t - R_t^{SE} \quad \text{(Eq 24)}$$

where $U_t$ is incoming precipitation, $I_t$ is infiltration into the soil column, and $R_t^{SE}$ is saturation-excess runoff generated when the ponding storage exceeds its capacity.

[36] In the PND configuration, infiltration depends on the total water available at the surface, consisting of both precipitation and previously ponded water:

$$I_t = (S_t^{PND} + U_t)G_t^U \qquad \text{(Eq 25)}$$

This formulation ensures that infiltration decreases as the soil approaches saturation, consistent with physical expectations.

[37] When surface storage exceeds the maximum ponding capacity $S_{max}^{PND}$, the excess is routed to saturation-excess runoff. This mechanism allows the model to represent delayed runoff generation while preserving mass conservation.

[38] Water infiltrating from the ponding layer contributes to soil storage according to:

$$\theta_{t+1} = \theta_t + I_t - O_t - L_t \qquad \text{(Eq 26)}$$

while the baseflow and evapotranspiration formulations remain unchanged from the NP configuration. Thus, the PND scenario introduces an intermediate surface storage that regulates how precipitation enters the soil column, allowing the model to represent short-term surface water accumulation and delayed infiltration without altering the internal soil-process structure.

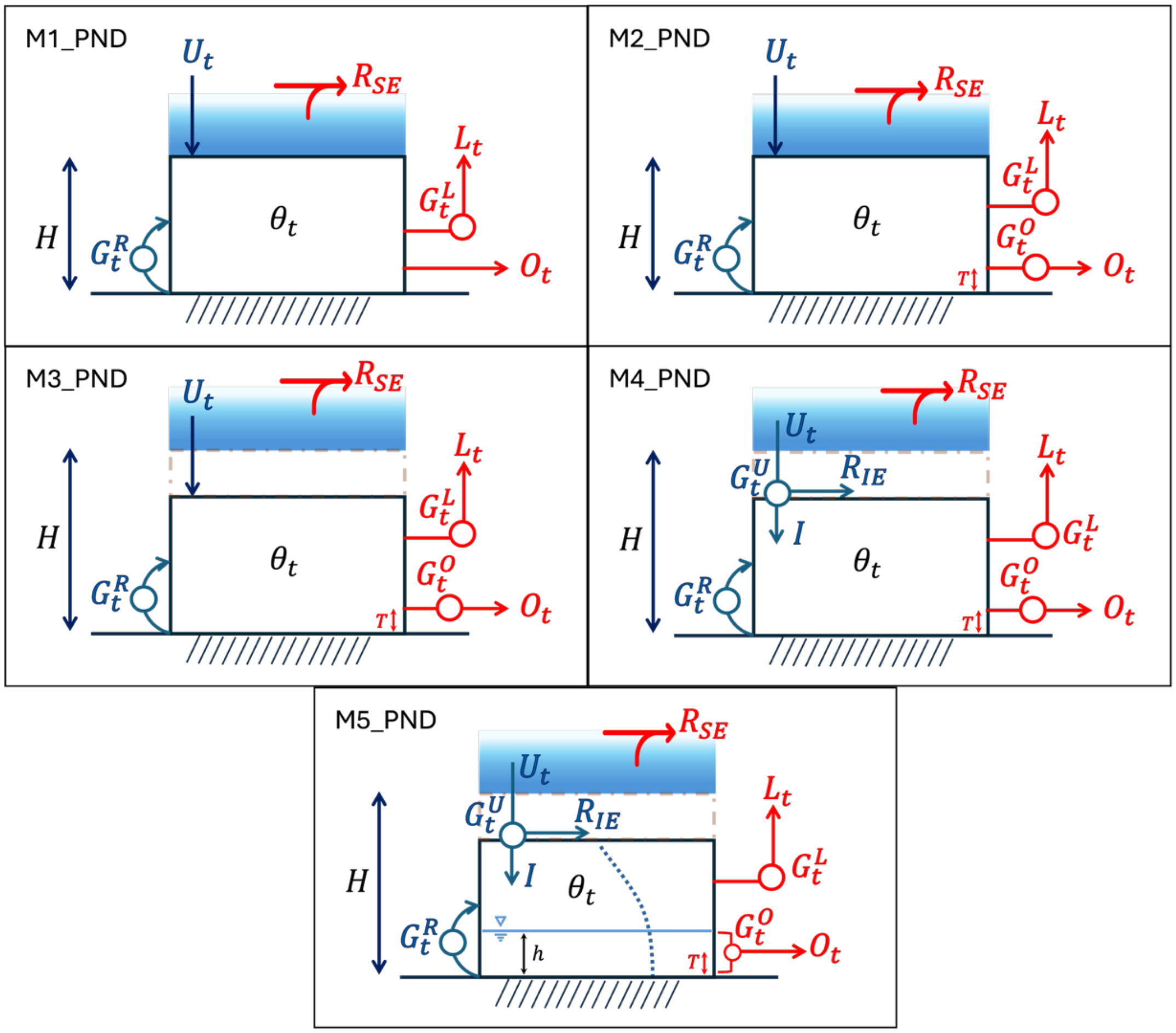

**Figure 4 Model configurations including surface ponding (PND).** Precipitation first accumulates in a temporary surface storage before infiltrating into the soil column or generating saturation-excess runoff. The five panels correspond to progressively

augmented MCP models (M1–M5) that incorporate bounded soil storage, state-dependent baseflow, variable porosity, infiltration capacity, and nonlinear water-table dynamics.

### 2.4 Representation of Vertical Drainage (DR)

[39] To represent deep percolation below the modeled soil column, we introduce a free vertical drainage mechanism referred to as the DR configuration. An additional downward flux $V_t$ removes water from the modeled soil storage and represents losses to deeper subsurface layers or recharge to groundwater storage outside the modeled system (**Figure 5**).

[40] The soil water balance is modified as:

$$\theta_{t+1} = \theta_t + U_t - O_t - R_t^{SE} - V_t - L_t \quad \text{(Eq 27)}$$

where $V_t$ is the vertical drainage flux, defined as:

$$V_t = G_t^V \theta_a \quad \text{(Eq 28)}$$

and $G_t^V$ is the vertical drainage gate, parameterized as:

$$G_t^V = \sigma(a_V \tilde{s}_t - b_V) \quad \text{(Eq 29)}$$

where $a_V$ and $b_V$ are learnable parameters. This formulation allows the model to learn how vertical drainage varies with soil moisture state, with drainage generally increasing as saturation increases.

[41] The DR configuration does not alter the internal process formulations for baseflow, evapotranspiration, or surface runoff; it adds an additional lower-boundary loss pathway through which water can leave the soil column. Applying this mechanism to the previously defined model hierarchy yields the corresponding drainage-enabled models M1_DR to M5_DR, allowing investigation of how deep percolation influences streamflow prediction across hydroclimatic regimes.

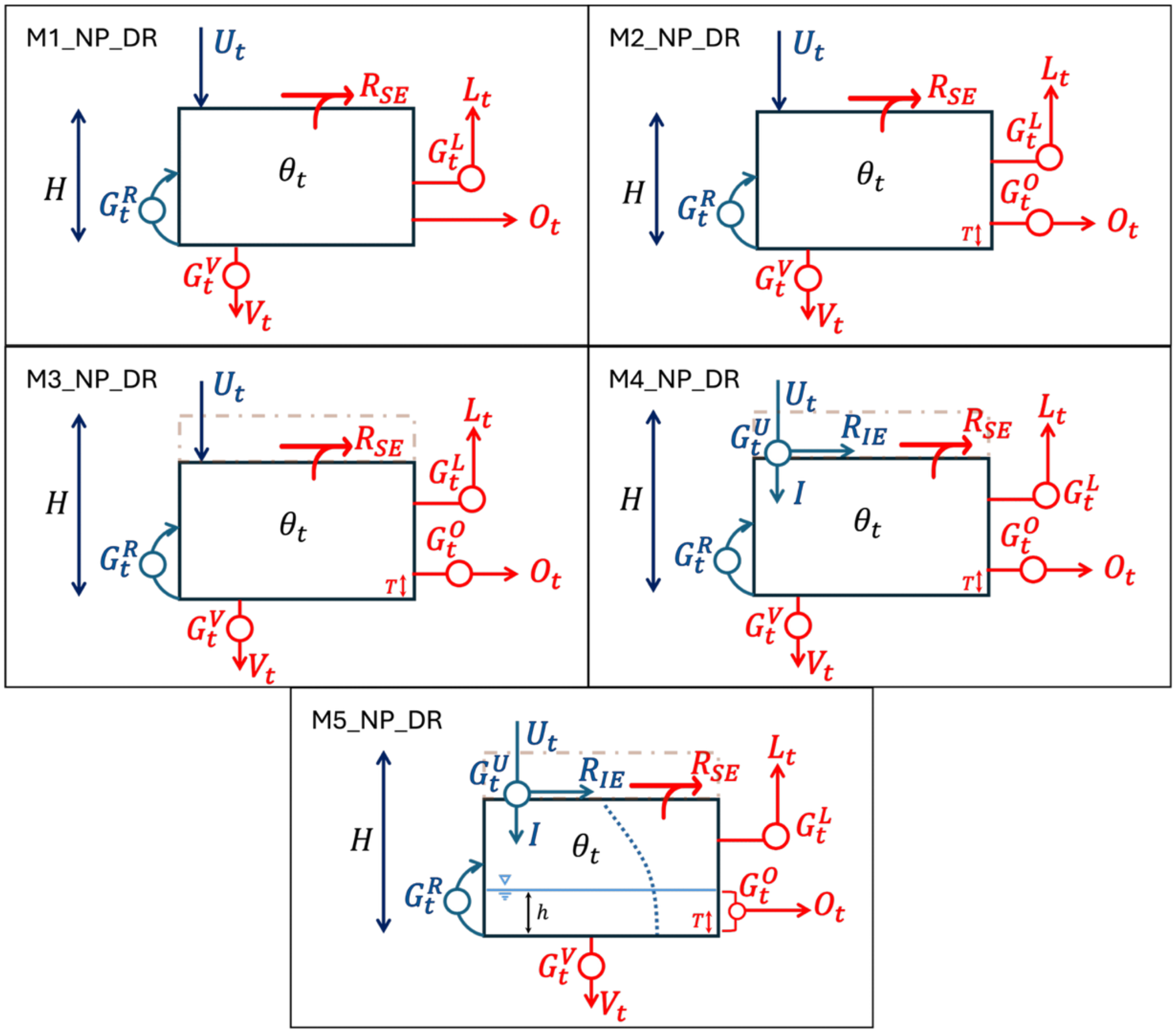

**Figure 5 Model configurations including free vertical drainage (DR) with no ponding.** An additional downward flux $V_t$ represents water draining from the bottom of the soil column to deeper subsurface storage. The drainage flux is controlled by a vertical drainage gate $G_t^V$ and depends on the available soil water above the wilting threshold. The panels illustrate the DR extension for the five progressively augmented MCP models (M1–M5). Blue elements denote storage and inflow processes, while red elements represent outgoing fluxes including baseflow ($O_t$), evapotranspiration ($L_t$), saturation-excess runoff ($R_t^{SE}$), infiltration-excess runoff ($R_t^{IE}$), and vertical drainage ($V_t$).

## 2.5 Model Scenario Combinations and Experimental Design

[42] The process representations described above define a hierarchy of internal soil-process models (M1–M5) and a set of alternative boundary-condition scenarios controlling how water enters and exits the soil column. Together, these dimensions define the full experimental space explored in this study.

[43] The internal soil-process hierarchy progressively augments the MCP storage unit with additional hydrological mechanisms, beginning with a bounded soil bucket representation (M1) and culminating in a formulation that includes state-dependent conductivity, variable soil porosity, infiltration capacity, and nonlinear water-table dynamics (M5). Each successive model therefore increases the level of physical realism embedded within the MCP storage unit while maintaining the same computational structure.

[44] In addition to internal process representations, four boundary-condition scenarios are considered that determine how water interacts with the soil column:

- **NP (No Ponding, No Drainage):** Precipitation interacts directly with the soil storage and no vertical drainage is allowed.
- **PND (Surface Ponding):** A temporary surface storage layer is introduced, allowing precipitation to accumulate before infiltrating into the soil column or producing surface runoff.
- **NP–DR (Free Vertical Drainage):** An additional downward flux allows water to drain from the bottom of the soil column, representing deep percolation beyond the modeled soil layer.
- **PND–DR (Combined Configuration):** Both surface ponding and vertical drainage are activated simultaneously.

[45] Combining the five internal soil-process models with the four boundary-condition scenarios results in a total of 20 model configurations evaluated in this study:

$$5 \text{ model structures } \times \text{ } 4 \text{ scenarios} = 20 \text{ models}$$

This structured experimental design enables systematic evaluation of how progressively-increasing physical realism influences model behavior and predictive performance across diverse hydroclimatic environments. Specifically, the framework allows assessment of the:

(i) Impact of progressively embedding hydrological processes within a single MCP storage unit

(ii) Influence of surface boundary conditions on infiltration and runoff generation, and

(iii) Role of vertical drainage in regulating soil water dynamics and streamflow response.

**Table 3. Experimental model configurations resulting from combinations of the five internal soil-process models and four boundary-condition scenarios.**

| Soil Model | NP | PND | NP-DR | PND-DR |
|---|---|---|---|---|
| M1 | M1_NP | M1_PND | M1_NP-DR | M1_PND-DR |
| M2 | M2_NP | M2_PND | M2_NP-DR | M2_PND-DR |
| M3 | M3_NP | M3_PND | M3_NP-DR | M3_PND-DR |
| M4 | M4_NP | M4_PND | M4_NP-DR | M4_PND-DR |
| M5 | M5_NP | M5_PND | M5_NP-DR | M5_PND-DR |

## 2.6 Study area

[46] All experiments in this study were conducted using data from 15 catchments selected from the CAMELS (Catchment Attributes and Meteorology for Large-sample Studies) dataset (*Addor et al., 2017*). CAMELS provides meteorological forcings, streamflow observations, and catchment attributes for hundreds of basins across the contiguous United States and has become a widely used benchmark dataset for large-sample hydrological modeling.

[47] The selected catchments span five major hydroclimatic regions, including arid, snow-dominated, and rainfall-dominated regimes across both western and eastern portions of the United States. These basins were chosen to represent a broad range of climatic and hydrological conditions, allowing the proposed modeling framework to be evaluated across diverse environmental settings (**Figure 6**).

[48] For each catchment, the dataset provides daily time series of mean areal precipitation (P, mm day$^{-1}$), potential evapotranspiration (PET, mm day$^{-1}$), and observed streamflow (Q, mm day$^{-1}$). The available data span the period 1980–2014. Following common practice in hydrological modeling studies, the time series were divided into three periods:

- Training period: 1987–2004
- Validation period: 1980–1987
- Testing period: 2004–2014

A three-year spin-up period was applied at the beginning of each simulation to minimize the influence of initial storage conditions. Model performance is evaluated exclusively on the test period, which is not used during model calibration or validation, thereby providing an independent assessment of predictive performance.

[49] In the simulations, total streamflow is computed as the sum of surface runoff and baseflow generated by the model storage unit. Because the focus of this study is on the internal hydrological processes represented within the storage unit, no routing scheme is applied between the modeled runoff components and the observed discharge.

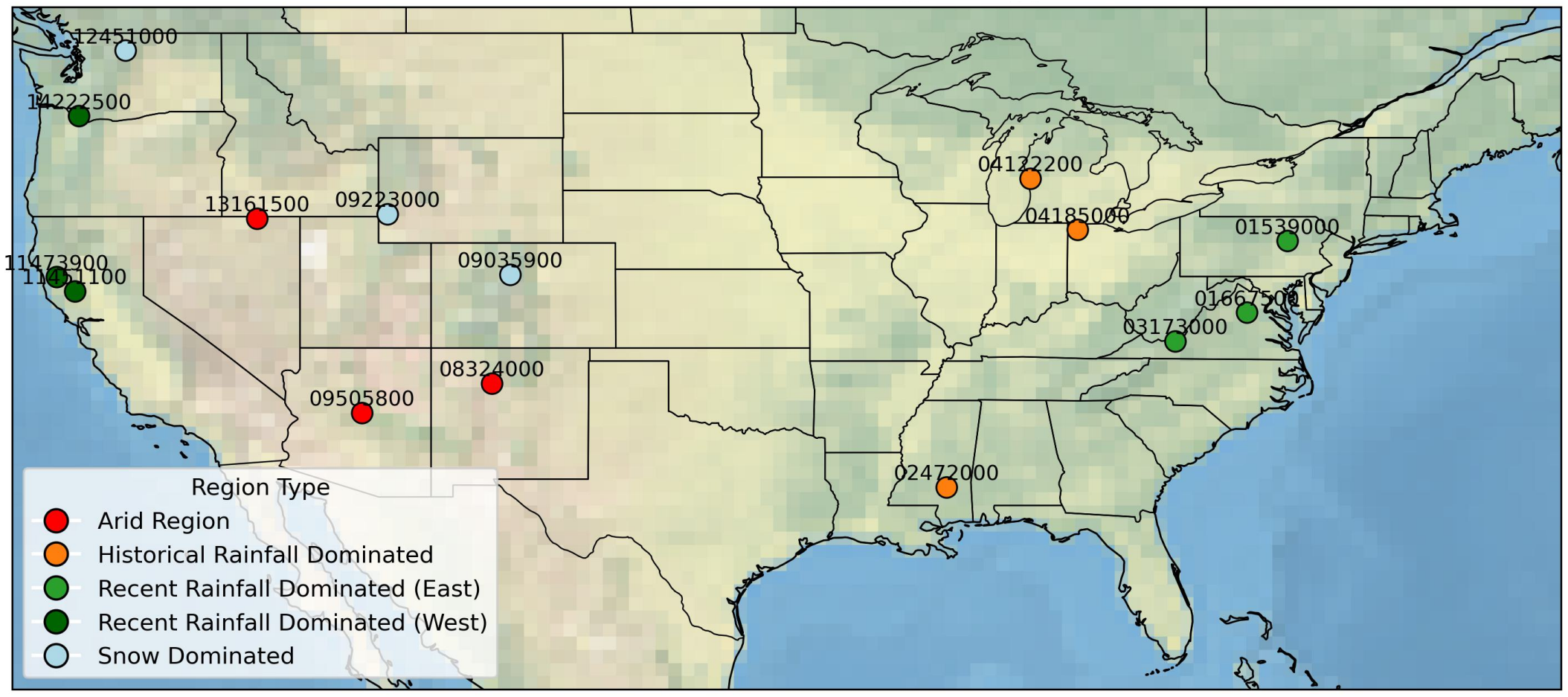

**Figure 6. Locations of the 15 selected CAMELS catchments representing five major hydroclimatic regions across the contiguous United States.**

## 2.7 Training Procedure

[50] Model parameters were optimized using gradient-based learning with the Adam optimizer. Each model was trained for 1000 epochs, although convergence was typically achieved after approximately 500 epochs. To account for sensitivity to initialization and learning-rate selection, the training procedure was conducted in two stages:

**(i) Initial parameter exploration:** Each model was first trained using 30 different random seeds. During this stage, the learning rate was set to 0.1, and a dynamic learning-rate schedule was applied using a learning-rate decay factor of 0.75. Model performance on the validation period was used to select the best parameter set among the 30 runs based on the Kling–Gupta Efficiency score (KGE) (*Gupta et al., 2009*).

**(ii) Learning-rate refinement:** The selected parameter set for each catchment and model was then used as the initial condition for a second training phase. In this stage, 10 additional simulations were conducted using different learning rates: $[\mathbf{0.001}, \mathbf{0.01}, \mathbf{0.1}, \mathbf{0.2}, \mathbf{0.3}, \mathbf{0.4}, \mathbf{0.5}, \mathbf{0.6}, \mathbf{0.7}, \mathbf{0.8}]$

Combining the initial exploration and refinement phases resulted in 40 training runs per catchment and model configuration. The parameter set yielding the highest validation KGE score was selected as the final calibrated model for that catchment.

## 2.8 LSTM Benchmark Model

[51] To provide a data-driven benchmark for comparison with the proposed MCP-based models, we trained a *Long Short-Term Memory* (LSTM) neural network to predict streamflow using the same meteorological inputs as the MCP models. LSTM networks have demonstrated strong predictive performance in rainfall–runoff modeling and are widely used as benchmark models in hydrological machine-learning studies.

[52] The LSTM model uses daily precipitation P (mm day$^{-1}$) and potential evapotranspiration PET (mm day$^{-1}$) as inputs and predicts daily streamflow Q (mm day$^{-1}$). To capture temporal dependencies in hydrological processes, the model was trained using input sequences of varying length. Following the approach used in previous studies, four sequence lengths were considered: 300, 330, 365, and 390 days. For each sequence length, the model was trained using 10 different random seeds, resulting in multiple realizations of the network initialization. Among these runs, the model achieving the highest validation performance was selected for each catchment. Training was performed using the Adam optimizer with learning rates of 0.01 and 0.03, a batch size of 64, and a maximum of 2000 training epochs. The KGE metric was used as the loss function during training to ensure consistency with the calibration objective used for the MCP models.

[53] Consistent with the MCP training framework, local models were trained independently for each catchment rather than using a regional model across basins. The same data partitions used for the MCP models were applied for the LSTM benchmark: 1987–2004 for training, 1980–1987 for validation, and 2004–2014 for testing. A three-year spin-up period was also applied at the beginning of each simulation to reduce the influence of initial conditions. This benchmark configuration allows direct comparison between the proposed MCP-based models and a state-of-the-art recurrent neural network trained using the same inputs, evaluation metrics, and experimental setup.

## 2.9 Evaluation Metrics

[54] Model performance was evaluated using the *Kling–Gupta Efficiency* (KGE) metric and its component statistics. KGE provides a comprehensive measure of hydrological model performance by simultaneously evaluating correlation, bias, and variability between simulated and observed streamflow.

[55] The KGE metric is defined as:

$$KGE = 1 - \sqrt{(1-\alpha)^2 + (1-\beta)^2 + (1-\rho)^2} \quad \text{(Eq 30)}$$

$$\alpha = \frac{\sigma_s}{\sigma_o} \quad \text{(Eq 31)}$$

$$\beta = \frac{\mu_s}{\mu_o} \qquad \text{(Eq 32)}$$

$$\rho = \frac{Cov_{so}}{\sigma_s \sigma_o} \qquad \text{(Eq 33)}$$

In these expressions: $\sigma_s$ and $\sigma_o$ are the standard deviations of the simulated and observed streamflow time series, respectively; $\mu_s$ and $\mu_o$ are the corresponding means; and $Cov_{so}$ denotes the covariance between simulated and observed streamflow. The three components of KGE represent distinct aspects of model performance: correlation ($\rho$) measures agreement in timing and pattern, variability ratio ($\alpha$) evaluates whether simulated variability matches observations, and bias ratio ($\beta$) indicates systematic over- or under-prediction of streamflow. KGE is maximized when all three components equal unity.

[56] In this study, we report the KGE skill score ($KGE_{ss}$), which provides a normalized version of KGE relative to a benchmark model, Following the formulation discussed by *Knoben et al. (2019)*:

$$KGE_{ss} = 1 - \frac{1 - KGE}{\sqrt{2}} \qquad \text{(Eq 34)}$$

Using this formulation ensures that $KGE_{ss} \leq 0$ when the model performs no better than a trivial benchmark model corresponding to simulated values at every time step being set to the observed long-term-mean, while a perfect model yields $KGE_{ss} = 1$. All reported evaluation metrics are computed over the independent test period (2004–2014).

# 3 Results

## 3.1 Overall Model Performance Across Catchments

[57] Test period predictive performance of the different model configurations was evaluated across the 15 study catchments using the $KGE_{ss}$ metric. **Figure 7** summarizes the distributions of $KGE_{ss}$ values across basins for five model structures (M1–M5) under four boundary-condition scenarios: standard without drainage (NP), ponding without drainage (PND), standard with drainage (NP_DR), and ponding with drainage (PND_DR). The dashed horizontal line indicates median performance of the LSTM benchmark model. Basin-by-basin performance across all configurations is shown in **Figure S1** of the supplementary materials.

[58] Across all scenarios, model performance improved systematically as model structural complexity increased from M1 to M5. The simplest configuration (M1) produced the lowest median skill and exhibited the largest inter-basin variability. Under the NP configuration, median $KGE_{ss}$ increased from ~0.62 for M1 to ~0.77 for M5 (**Table S1** in the supplementary materials), indicating a substantial improvement in predictive skill as additional process representations were incorporated. A similar trend is observed for the PND scenario, where the median skill increases from ~0.62 to ~0.76 from M1 to M5. These results suggest that the progressive incorporation of additional hydrological processes enables better representation of dominant runoff-generation mechanisms across diverse hydroclimatic regimes.

[59] The basin-level patterns shown in **Figure S1** further highlight this structural improvement. Across most catchments, higher-order model structures (M4–M5) consistently achieve higher $KGE_{ss}$ values than simpler configurations (M1–M2). This pattern indicates that the additional process representations embedded in the higher-order MCP structures improve the model's ability to reproduce observed hydrological behavior across diverse hydroclimatic settings. However, in some basins the addition of further process complexity does not necessarily improve predictive performance, indicating that dominant hydrological processes vary across catchments.

[60] Among the tested boundary-condition scenarios, *inclusion of vertical drainage produced the most consistent improvements in predictive skill*. Enabling drainage increased the median $KGE_{ss}$ of the simplest

model (M1) from ~0.62 to ~0.73, representing a skill improvement of more than 0.10. Drainage-enabled configurations also reduced the spread of performance across basins, indicating improved robustness across hydroclimatic conditions. The highest-performing MCP configuration, M5_NP_DR, achieved a median $KGE_{ss}$ of ~0.77 across the study catchments.

[61] In contrast, the inclusion of ponding processes produced relatively modest changes in predictive skill. Differences between NP and PND configurations were generally small across model structures, suggesting that ponding processes may have a secondary influence on model performance compared with subsurface drainage processes in the studied catchments.

[62] As expected, the data-based *LSTM benchmark achieved the highest overall median skill*, with a median $KGE_{ss}$ of ~0.80 across basins. However, the best-performing MCP configurations approached this benchmark closely. For example, the M5_NP_DR configuration achieved a median $KGE_{ss}$ only ~0.03 lower than the LSTM, indicating that physically-interpretable MCP-based models can achieve predictive performance comparable to state-of-the-art data-driven approaches while maintaining explicit representation of hydrological processes.

[63] Overall, these results demonstrate that both *model structural complexity and the inclusion of subsurface drainage processes play important roles in improving predictive performance* across diverse catchments. While the LSTM benchmark provides slightly higher median skill, the competitive performance of the MCP configurations highlights the potential of physically-interpretable models to bridge the gap between process-based and purely data-driven hydrological modeling approaches.

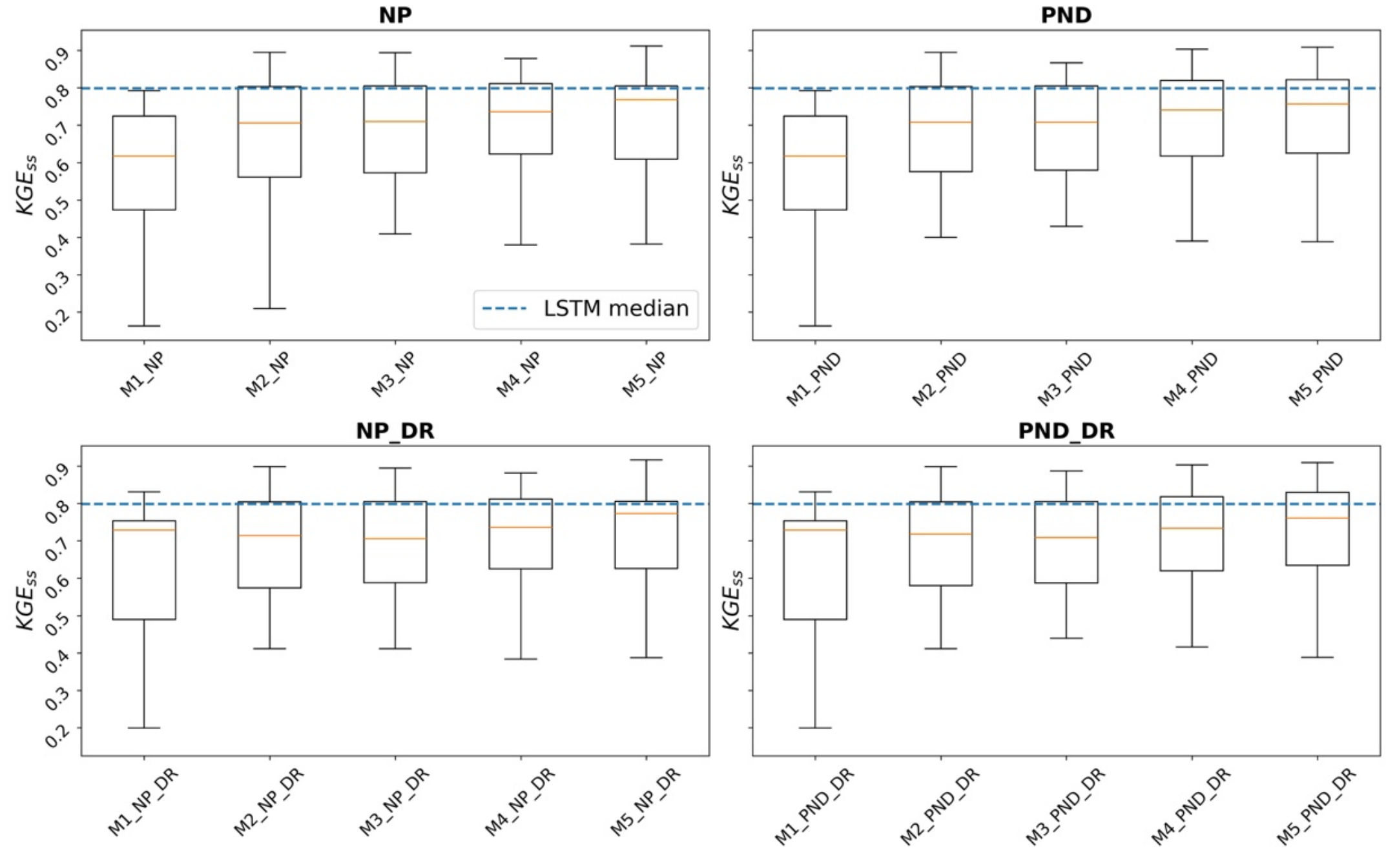

**Figure 7. Distribution of predictive performance across the 15 study catchments measured using the $KGE_{ss}$.** Boxplots show the performance of five model structures (M1–M5) under four boundary-condition scenarios: standard without drainage (NP), ponding without drainage (PND), standard with drainage (NP_DR), and ponding with drainage (PND_DR). The dashed horizontal line represents the median performance of the LSTM benchmark model. Increasing model structural complexity generally improves predictive skill across basins, while the inclusion of vertical drainage processes leads to substantial

performance gains. Ponding processes produce comparatively smaller changes in model performance.

**Table 4. Basin-level comparison of predictive performance between the best-performing MCP configuration and the LSTM benchmark across the 15 study catchments.** For each basin, the MCP model configuration with the highest $KGE_{ss}$ is reported along with the corresponding LSTM performance. The table provides the numerical basis for the comparison shown in Figure 9 and illustrates that MCP models outperform the LSTM benchmark in 7 of the 15 catchments.

| Basin | Model | Best $KGE_{ss}$ | LSTM $KGE_{ss}$ |
|---|---|---|---|
| Arid Region (08324000) | M5_PND_DR | **0.88** | **0.79** |
| Arid Region (09505800) | M4_PND_DR | 0.79 | 0.83 |
| Arid Region (13161500) | M5_NP | 0.70 | 0.77 |
| Historical Rainfall Dominated (02472000) | M4_PND | **0.88** | **0.74** |
| Historical Rainfall Dominated (04122200) | M5_PND | 0.80 | 0.91 |
| Historical Rainfall Dominated (04185000) | M5_DR | **0.92** | **0.74** |
| Recent Rainfall Dominated (East) (01539000) | M5_PND | 0.88 | 0.91 |
| Recent Rainfall Dominated (East) (01667500) | M5_PND_DR | **0.85** | **0.81** |
| Recent Rainfall Dominated (East) (03173000) | M5_NP | 0.71 | 0.76 |
| Recent Rainfall Dominated (West) (11451100) | M5_NP | **0.91** | **0.86** |
| Recent Rainfall Dominated (West) (11473900) | M5_DR | **0.88** | **0.80** |
| Recent Rainfall Dominated (West) (14222500) | M4_NP | 0.80 | 0.93 |
| Snow Dominated (09035900) | M4_PND_DR | 0.29 | 0.69 |
| Snow Dominated (09223000) | M1_DR | **0.81** | **0.80** |
| Snow Dominated (12451000) | M5_PND_DR | 0.71 | 0.76 |

### 3.2 Comparison with LSTM Benchmark

[64] To further evaluate predictive capability of the MCP framework relative to data-driven approaches, model performance was compared against the LSTM benchmark across all 15 study catchments. **Figure 8** presents a basin-level $KGE_{ss}$ comparison between the *best-performing* MCP configuration for each catchment and the LSTM model. Each point represents one catchment, and points above the 1:1 line indicate cases where the MCP model outperforms the LSTM benchmark.

[65] Overall, the best-performing MCP configurations achieved slightly higher predictive skill across the full set of catchments, with a median $KGE_{ss}$~0.81, compared with ~0.80 for the LSTM (**Table S4**). Moreover, the basin-level comparison reveals that MCP models achieved higher skill than the LSTM benchmark in *7 out of the 15 catchments* (**Figure 8; Table 4**). These results demonstrate that physically-interpretable MCP models can achieve performance comparable to state-of-the-art deep learning approaches in many hydroclimatic settings.

[66] The basin-specific comparison further highlights how relative performance of the two modeling approaches varies across hydroclimatic regimes. For example, in several rainfall-dominated basins, MCP configurations achieved higher skill than the LSTM benchmark. In the *Historical Rainfall Dominated basin* (04185000), the MCP configuration M5_DR achieved a $KGE_{ss} =$ 0.92, substantially exceeding the LSTM performance of 0.74. Similarly, in the *Recent Rainfall Dominated (West) basin* (11451100), the MCP configuration M5_NP achieved a $KGE_{ss} =$ **0.91**, compared with 0.86 for the LSTM model. These results suggest that the physically-interpretable process representations embedded within the MCP framework can effectively capture dominant hydrological dynamics in rainfall-dominated environments.

[67] In contrast, the LSTM model exhibited stronger performance in some catchments, particularly in snow-dominated regions. For example, in the *Snow Dominated basin* (09035900), the LSTM achieved a $KGE_{ss} =$ 0.69, whereas the best-performing MCP configuration achieved a substantially lower value (only 0.29). In the current MCP formulation, a single storage unit represents total water storage without explicitly distinguishing snowpack. However, snow-dominated systems are strongly controlled by accumulation and delayed release of snow water. Prior work (e.g., *Wang et al., 2025*) shows that representing these processes typically requires a *dedicated snow storage* (SWE) state. The absence of such a component likely limits MCP performance, whereas the LSTM can implicitly learn these dynamics.

[68] A distributional comparison of model skill further illustrates the differences between MCP configurations and the LSTM benchmark. **Table S2** summarizes the percentile statistics of $KGE_{ss}$ across catchments for each model configuration. The LSTM model exhibits consistently better lower-percentile performance, with a *5th percentile* $KGE_{ss}$~ *0.72*, whereas MCP configurations show worse minimum performance, with worst-case $KGE_{ss}$ values ranging from ~0.16 to ~0.28. This indicates that the LSTM model provides more consistent performance across catchments, particularly under challenging hydrological settings.

[69] However, the upper percentiles of the MCP configurations approach those of the LSTM model. For example, the *95th percentile* $KGE_{ss}$ of the best MCP configurations reaches ~0.89, close to 0.92 for the LSTM benchmark. This suggests that when the MCP framework successfully captures the dominant hydrological processes within a catchment, it can achieve predictive skill comparable to data-driven approaches.

[70] Overall, these results indicate that while the LSTM benchmark maintains slightly *better and more consistent predictive performance across catchments*, the MCP framework demonstrates *competitive skill in many basins and in several cases exceeds the performance of the deep learning model*. Importantly, the MCP framework achieves this performance while preserving *physical interpretability and explicit*

*representation of hydrological processes*, providing a promising pathway toward bridging the gap between process-based and data-driven hydrological modeling.

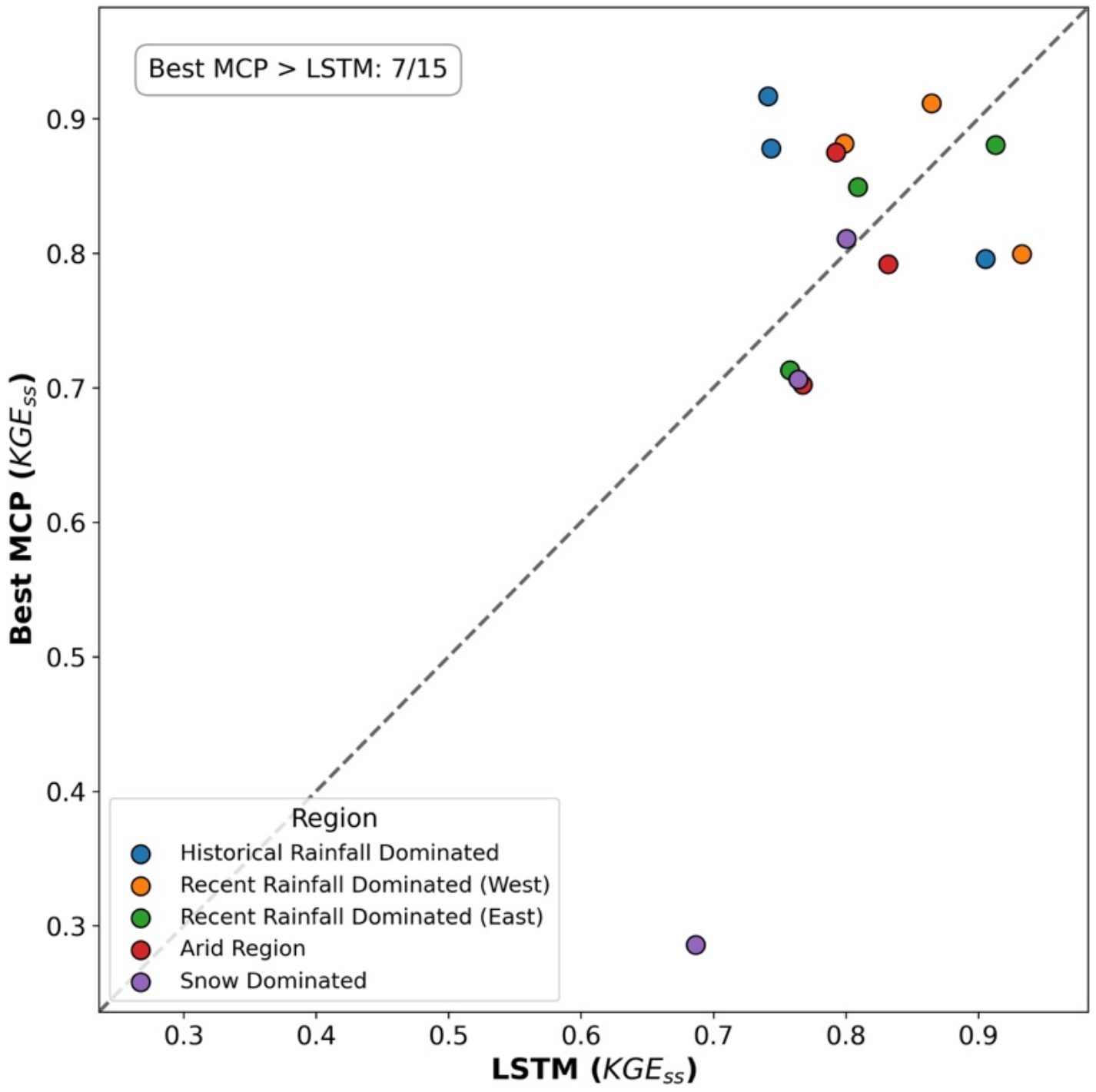

**Figure 8. Comparison of predictive performance between the best-performing MCP configuration and the LSTM benchmark across the 15 study catchments.** Each point represents one basin and is colored by hydroclimatic region. The dashed line indicates the 1:1 relationship between MCP and LSTM performance. Points above the line represent basins where the MCP model outperforms the LSTM benchmark. MCP configurations achieve higher $KGE_{ss}$ values than LSTM in **7 of the 15 catchments**, demonstrating the competitive performance of the physically-interpretable MCP framework relative to the deep learning benchmark.

### 3.3 Impact of Progressive Physical Augmentation (M1 → M5)

[71] To isolate the effects of progressively-increasing model structural complexity, we quantified improvements in predictive skill relative to the baseline configuration (M1). **Figure 9** summarizes the distribution of incremental performance gains ($\Delta KGE_{ss}$) for models M2–M5 relative to M1 across the four boundary-condition scenarios (NP, PND, NP_DR, and PND_DR), and **Table S3** reports the values used in **Figure 9**. Positive values indicate improved predictive performance compared with the baseline M1 model.

[72] Across all scenarios, progressively augmenting the model structure resulted in systematic improvements in predictive skill. Under the standard configuration without drainage (NP), the median improvement increased from ~0.08 for M2 to ~0.09 for M4 (**Table S3**), while the most complex configuration (M5) produced a similar median improvement of ~0.08 relative to M1. Maximum improvements reached 0.30 $\Delta KGE_{ss}$ in individual basins, indicating that structural augmentation can substantially enhance predictive performance in certain hydroclimatic settings. Similar patterns were

observed for the ponding configuration (PND), where median improvements ranged from 0.08–0.11 (**Table S3**), with the largest gains again observed for M4 and M5.

[73] When vertical drainage processes were enabled, the magnitude of incremental improvements relative to M1 became smaller but remained consistently positive. Under the NP_DR scenario, median improvements ranged from 0.05 (M2) to 0.08 (M5), and under PND_DR they ranged from 0.05–0.08 (**Table S3**). The reduced magnitude of $\Delta KGE_{ss}$ under drainage-enabled configurations reflects the fact that the baseline model (M1) already benefits substantially from the inclusion of subsurface drainage processes, leaving less room for additional gains from structural augmentation.

[74] Basin-level responses to increasing model complexity are shown in **Figure S2**, which illustrates the evolution of predictive performance across all 15 catchments as model structure progresses from M1 to M5. In most basins, the largest improvement occurs between M1 and M2, indicating that the introduction of *saturation-dependent horizontal drainage rate* yields the largest marginal gain in predictive skill. Subsequent increases in complexity (M3–M5) generally produce smaller but still positive improvements, suggesting diminishing returns as additional processes are incorporated.

[75] Overall, these results demonstrate that progressively augmenting the physical representation of hydrological processes leads to consistent improvements in predictive performance across diverse catchments. While the largest gains occur with the initial addition of key processes beyond the baseline model, higher-order configurations continue to provide incremental benefits, particularly in catchments where complex runoff-generation mechanisms are important

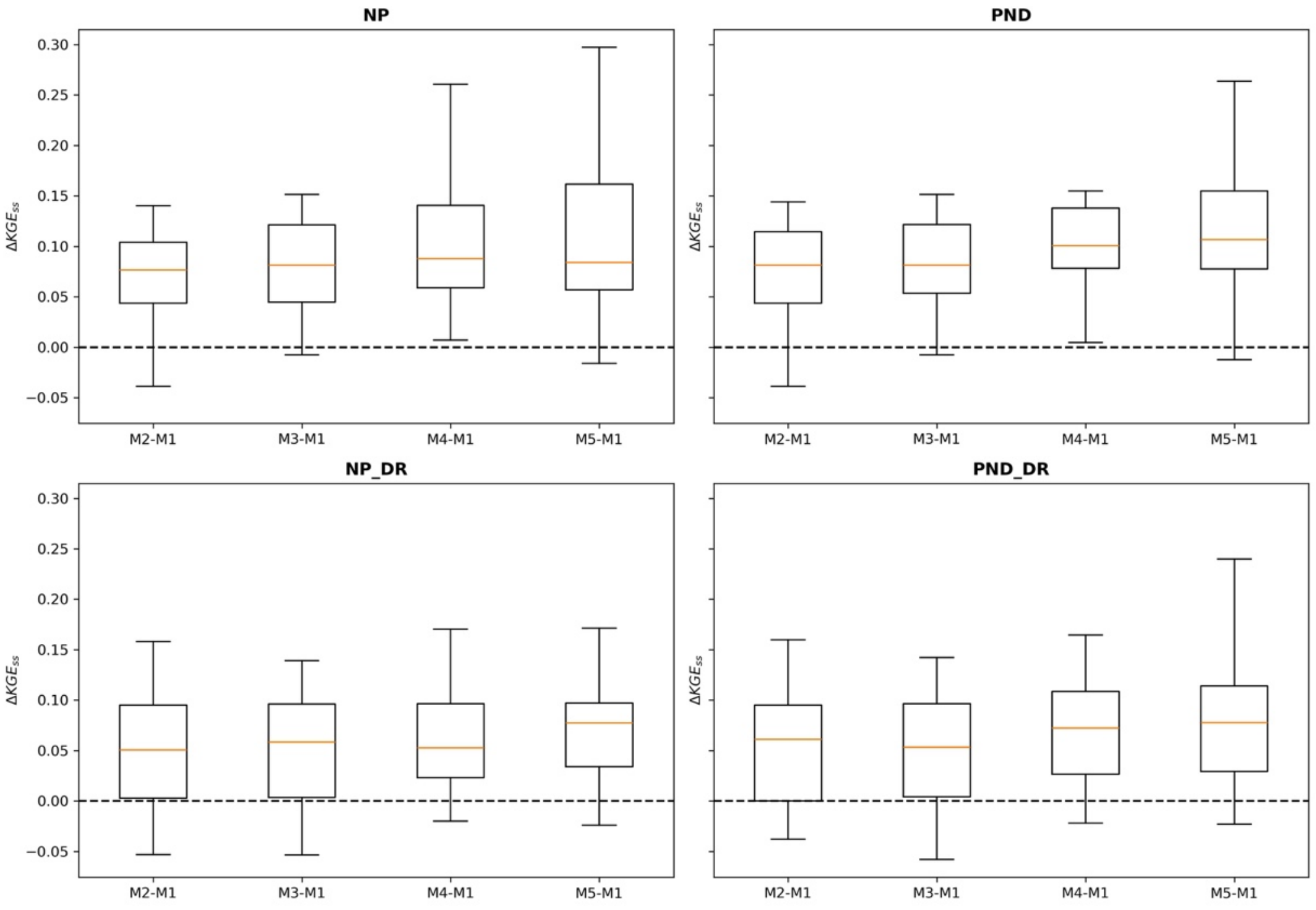

**Figure 9. Improvement in predictive skill relative to the baseline model (M1) measured as $\Delta KGE_{ss}$ across the 15 study catchments.** Boxplots show the distribution of performance gains for models M2–M5 relative to M1 under four boundary-condition scenarios (NP, PND, NP_DR, and PND_DR). Positive values indicate improved performance relative to the baseline configuration.

### 3.4 Influence of Boundary Conditions (NP, PND, DR, PND–DR)

[76] To evaluate how external boundary conditions influence predictive performance, we compared model skill under four configurations: no ponding and no drainage (NP), ponding without drainage (PND), drainage without ponding (NP_DR), and combined ponding and drainage (PND_DR). **Figure 10** shows the regional median response of $KGE_{ss}$ to increasing model complexity under these boundary-condition scenarios for representative hydroclimatic regimes, while **Table 5** summarizes the regional median differences between configurations. Additional results for the best-performing model structure in each region are provided in **Table S4** in the supplementary materials.

[77] *Arid Region:* The effect of boundary conditions is strongly region dependent. In the *arid region* (**Figure 10b**), activating vertical drainage substantially improves predictive performance across the model hierarchy. Averaged across model structures, drainage increases regional median $KGE_{ss}$ by ~0.17 for the NP configuration and ~0.15 for the PND configuration (**Table 5**). Although increasing structural complexity within the no-drainage family still improves skill, *the dominant shift in performance occurs when drainage is enabled*. This result suggests that allowing water to leave the modeled soil column is essential in arid systems, where deep percolation and delayed subsurface losses are important components of the catchment water balance. For the best-performing structure in this region (M4), adding drainage increases performance by ~0.06 under NP and ~0.07 under PND (**Table S4**), whereas ponding alone has only a small effect.

[78] *Snow-Dominated Region:* A similarly strong influence of drainage is observed in the *snow-dominated region* (**Figure 10c**). Averaged across model structures, activating drainage increases regional median $KGE_{ss}$ by ~0.38 for both NP and PND configurations (**Table 5**), far exceeding the effect of ponding alone. For the best-performing structure in this region (M5), drainage improves performance by ~0.41 under NP and ~0.42 under PND (**Table S4**). These large gains indicate that the lower boundary condition exerts a dominant control on model performance in snow-influenced basins. In these systems, drainage likely represents delayed release pathways and subsurface storage processes that become important during snow accumulation and melt periods.

[79] *Rainfall-Dominated Regions:* In contrast, the role of drainage is markedly different in *rainfall-dominated regions* (**Figure 10a**). In both the historical rainfall-dominated and recent rainfall-dominated regimes, enabling drainage generally reduces model performance. Across model structures, the mean effect of drainage ranges from approximately −0.05 in the historical rainfall-dominated region to approximately −0.08 in both recent rainfall-dominated regions. For the best-performing structures, drainage reduces performance by ~0.02 in the historical rainfall-dominated region, 0.10–0.12 in the recent rainfall-dominated East region, and 0.08 in the recent rainfall-dominated West region. These results indicate that *free drainage removes too much water from the active storage in humid rainfall-dominated systems*, thereby weakening the model's ability to sustain runoff generation and reproduce observed streamflow dynamics.

[80] Across all hydroclimatic regimes, the effect of *ponding* is consistently smaller than the effect of drainage. $KGE_{ss}$ differences between the NP and PND configurations are typically 0.001–0.01 (**Table 5**). Even in regions where ponding produces modest improvements, the magnitude of the effect remains much smaller than that of drainage, indicating that *surface storage processes play a secondary role relative to the lower boundary condition.*

[81] Taken together, these results demonstrate that the influence of boundary conditions is strongly hydroclimate dependent. Vertical drainage substantially improves performance in *arid and snow-dominated regions*, but generally degrades performance in *rainfall-dominated basins*. Ponding produces

comparatively modest changes across all regimes. These findings suggest that there is *no universally optimal boundary condition for all catchments*, and that appropriate representations of surface and subsurface boundary processes should be selected based on regional hydrological characteristics.

[82] These results highlight the importance of representing physically appropriate boundary processes within process-aware machine learning models, as incorrect boundary assumptions can substantially degrade predictive performance even when internal model structure is improved.

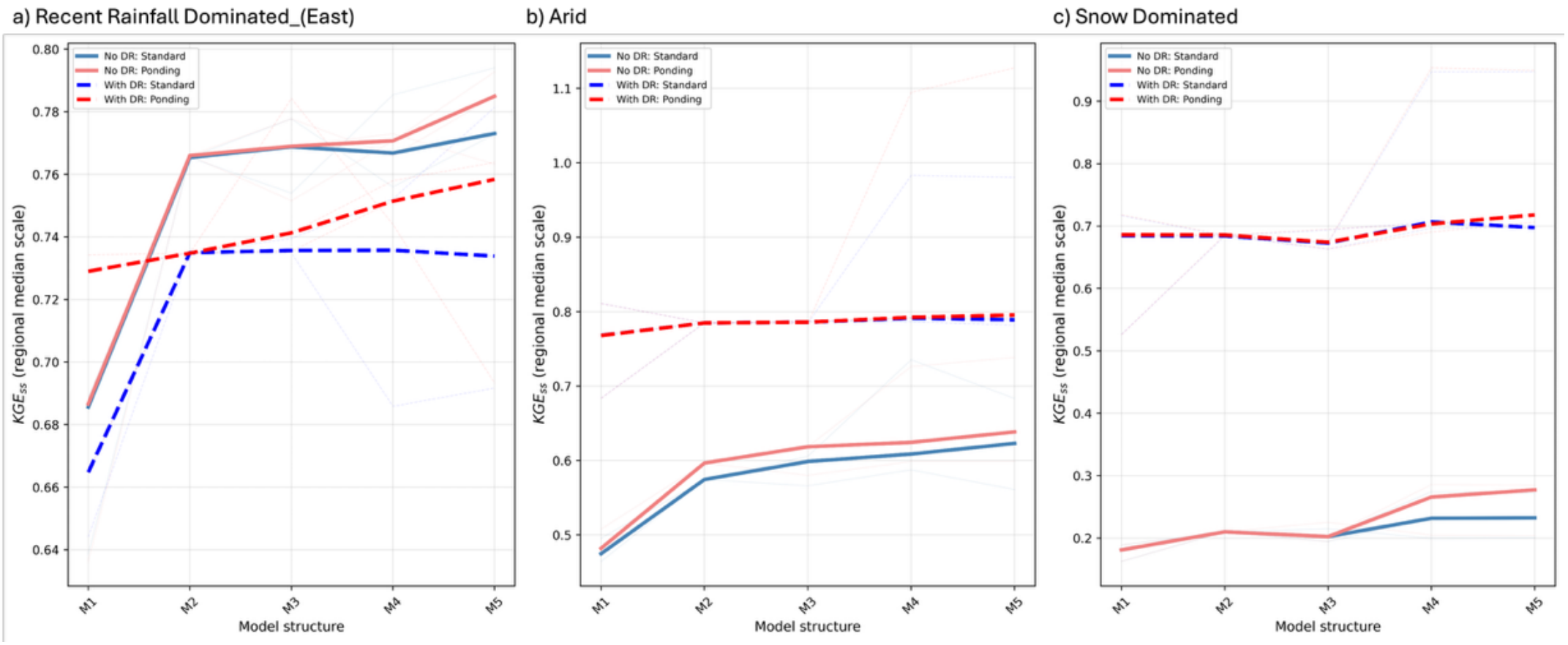

**Figure 10. Regional median predictive performance ($KGE_{SS}$) as a function of model structural complexity (M1–M5) under four boundary-condition scenarios: no ponding and no drainage (NP), ponding without drainage (PND), drainage without ponding (NP_DR), and combined ponding and drainage (PND_DR).** Panels show representative hydroclimatic regimes, illustrating how the influence of boundary conditions varies across climatic settings. In arid and snow-dominated regions, enabling vertical drainage substantially improves predictive skill, whereas in rainfall-dominated regions configurations without drainage generally perform better. Thin lines indicate individual basin responses and bold lines denote regional medians.

**Table 5. Regional median differences in predictive performance ($KGE_{ss}$) between boundary-condition configurations averaged across model structures (M1–M5).** Positive values indicate improved performance when the first configuration is used. The results highlight the dominant influence of vertical drainage relative to surface ponding across hydroclimatic regimes.

| Region | NP_DR − NP | PND_DR − PND | PND − NP | PND_DR − NP_DR | Ponding effect | Drainage effect |
|---|---|---|---|---|---|---|
| Arid Region | 0.17 | 0.15 | 0.004 | 0.004 | 0.004 | 0.16 |
| Historical Rainfall Dominated | −0.048 | −0.051 | 0.01 | -0.0004 | 0.01 | -0.05 |
| Recent Rainfall Dominated (East) | −0.080 | −0.081 | 0.001 | 0.004 | 0.001 | -0.08 |
| Recent Rainfall Dominated (West) | −0.079 | −0.080 | 0.0003 | 0.0004 | 0.0003 | -0.08 |
| Snow Dominated | 0.38 | 0.38 | 0.01 | 0.0007 | 0.01 | 0.38 |

### 3.5 Flow Regime Performance

[83] To better understand how model performance varies across the discharge spectrum, we evaluated predictive skill separately for *low-, mid-, and high-flow regimes*. Flow regimes were defined using exceedance probabilities, and model performance was assessed using both $RMSE$ and $KGE_{ss}$. **Figure 11** summarizes the distribution of errors and skill scores across flow regimes for the best-performing MCP configurations alongside the LSTM benchmark, while **Figure 12** presents representative flow duration curves (FDCs) for selected basins across different hydroclimatic regions.

[84] Across all models, error magnitude increases from low to high flows, reflecting the larger variability associated with peak discharge events. In the *low-flow regime*, the baseline configuration (M1) exhibits substantially larger RMSE and strongly negative $KGE_{ss}$ values, indicating poor reproduction of observed low-flow dynamics. Incorporating additional hydrological processes in the augmented MCP structures (M2–M5) substantially reduces RMSE and improves $KGE_{ss}$, demonstrating that these configurations better capture *slow storage dynamics and subsurface flow processes that control baseflow conditions*. Despite these improvements, the LSTM benchmark generally achieves the lowest RMSE and highest $KGE_{ss}$ in this regime, suggesting that the data-driven model more effectively reproduces the persistence and variability of low flows.

[85] In the *mid-flow regime*, both RMSE and $KGE_{ss}$ indicate improved performance relative to the baseline model. The difference between MCP configurations becomes smaller, with M2–M5 exhibiting broadly comparable error and skill distributions. This pattern suggests that once key hydrological processes are incorporated, *additional increases in model complexity yield only modest improvements in predictive performance*.

[86] In the *high-flow regime*, RMSE values increase for all models due to the larger magnitude and episodic nature of high-flow events. However, $KGE_{ss}$ values remain positive across models, indicating that the general timing and variability of peak flows are reasonably captured. Differences between MCP configurations are relatively small in this regime, although the LSTM model tends to maintain slightly higher skill scores across basins.

[87] To further evaluate model behavior across the full flow distribution, **Figure 12** presents flow duration curves for representative basins spanning arid, snow-dominated, and rainfall-dominated hydroclimatic regimes. These plots illustrate how well the models reproduce observed streamflow distributions across the entire range of exceedance probabilities.

[88] In the *arid basin* (**Figure 12a**), the best-performing MCP configuration closely follows the observed flow distribution across much of the exceedance spectrum but underestimates very low flows, resulting in a sharper drop in the tail of the FDC. The LSTM model captures low-flow persistence more effectively in this basin.

[89] In the *snow-dominated basin* (**Figure 12b**), the MCP model captures the overall slope of the observed FDC reasonably well but tends to overestimate moderate flows and underestimate extreme flows. The LSTM model again better reproduces the lower tail of the distribution. In the *rainfall-dominated basins* (**Figures 12c–12e**), both models reproduce the general shape of the FDC, although differences appear at both high- and low-exceedance probabilities. In several cases, the MCP configuration better

captures intermediate flows, whereas the LSTM model more accurately reproduces the persistence of low flows.

[90] Overall, the flow-regime analysis indicates that the largest improvements associated with progressive model augmentation occur in the *low-flow regime*, where additional physical process representations substantially reduce prediction errors relative to the baseline model. Differences between model structures become smaller in the *mid- and high-flow regimes*, suggesting diminishing returns from additional complexity once key runoff-generation mechanisms are represented. While the LSTM benchmark generally provides the most accurate representation of the full flow distribution, particularly for low-flow persistence, the MCP models capture many aspects of observed flow dynamics with comparable accuracy across moderate and high flows.

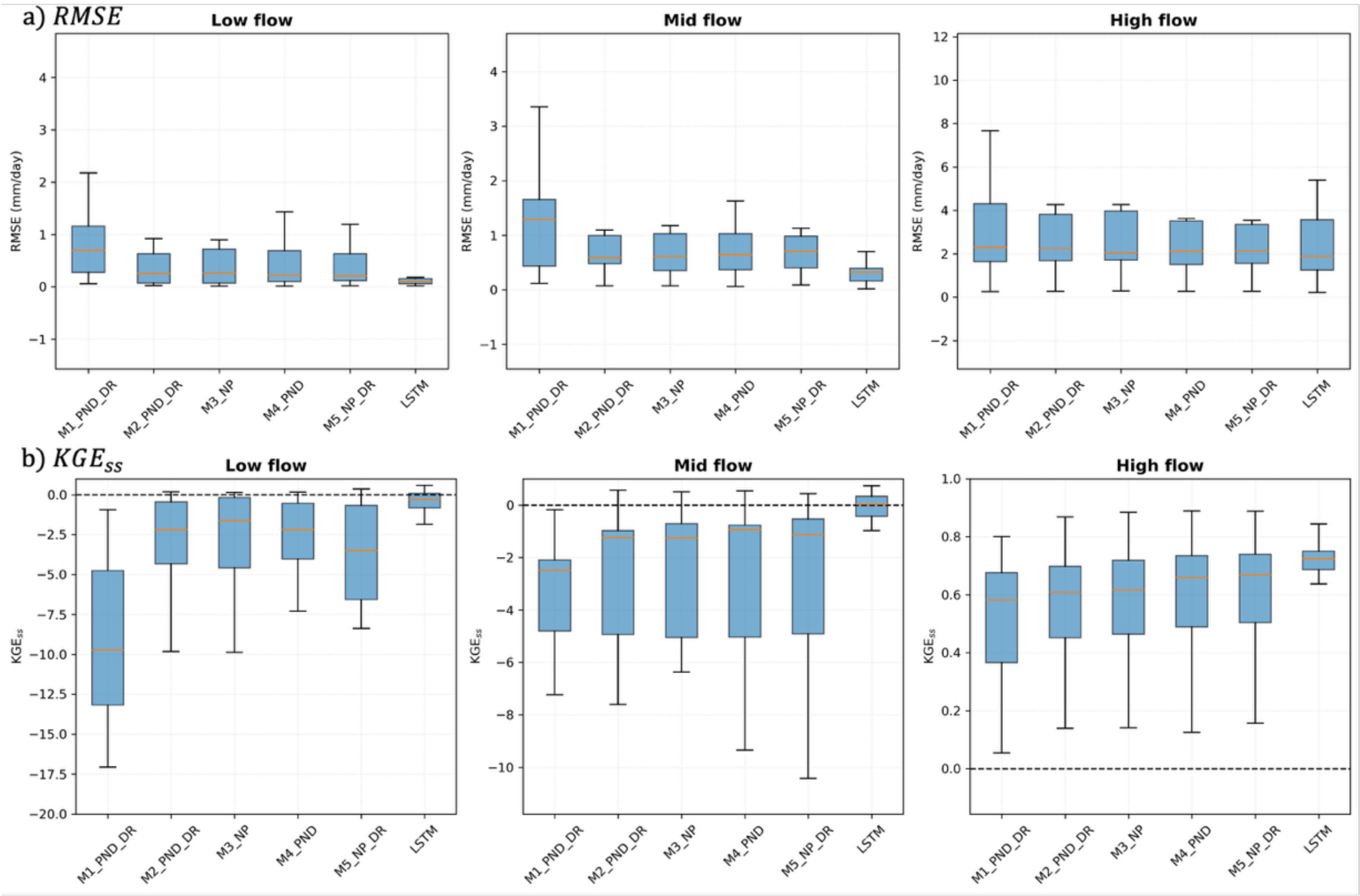

**Figure 11. Flow-regime performance across models.** Top row: RMSE distributions for low, mid-, and high-flow regimes. Bottom row: corresponding $KGE_{ss}$ distributions.

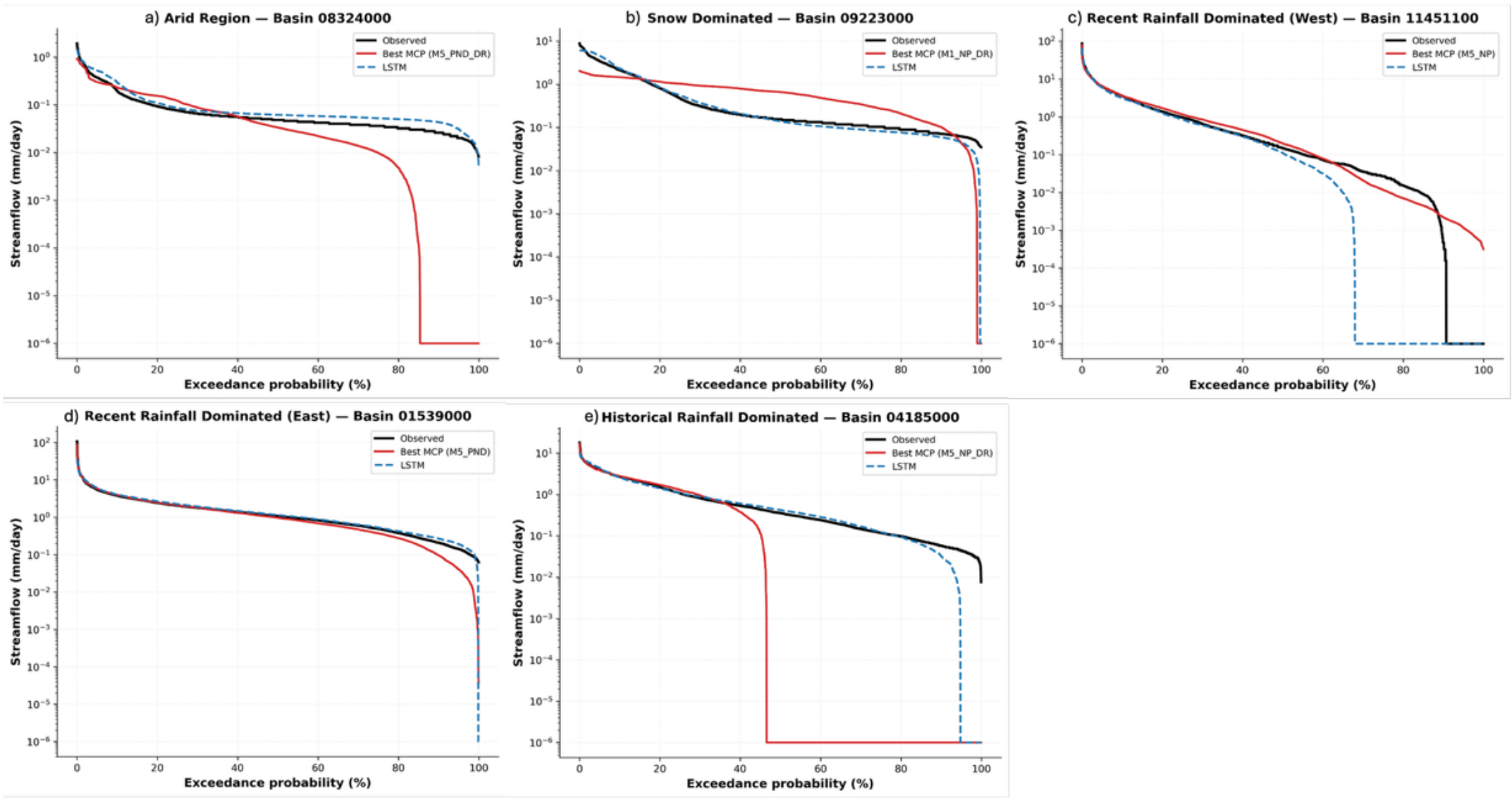

**Figure 12. Flow duration curves for representative basins across hydroclimatic regimes comparing observed streamflow, the best-performing MCP configuration, and the LSTM benchmark.**

### 3.6 Event-Scale Dynamics and Hydrograph Behavior

[91] Hydrograph comparisons provide additional insight into the temporal dynamics of simulated streamflow and help interpret the statistical performance metrics. While the flow-regime analysis (**Figure 11**) summarizes model skill across low-, mid-, and high-flow conditions and the flow duration curves (**Figure 12**) describe the distribution of simulated flows across exceedance probabilities, hydrographs provide a *time-resolved perspective* of how model differences emerge during individual runoff events. **Figure 13** illustrates representative hydrographs for five basins spanning the major hydroclimatic regimes considered in this study, including arid, snow-dominated, rainfall-dominated (West), rainfall-dominated (East), and historically rainfall-dominated systems. For each basin, three hydrological years representing dry, moderate, and wet conditions are shown.

[92] Across most basins, both the MCP models and the LSTM benchmark reproduce the general timing of major runoff events, indicating that both modeling approaches capture the dominant temporal variability of streamflow. Differences between the models primarily appear in the representation of peak magnitude, recession behavior, and low-flow conditions.

[93] In the *arid basin*, runoff events are episodic and driven by a small number of high-intensity precipitation events. Both models reproduce the timing of major runoff pulses with little delay relative to the observations. However, the LSTM model tends to generate sharper and more abrupt peak responses, particularly during wet years, whereas the MCP simulations produce smoother peaks and more gradual recessions. This smoother behavior reflects the influence of storage and drainage processes represented within the MCP framework, which distribute runoff generation over longer time scales.

[94] In the *snow-dominated basin*, streamflow exhibits a gradual seasonal rise associated with snow accumulation and subsequent melt. Both models capture the timing of the primary snowmelt peak, although differences occur in the magnitude of the peak and the representation of early-season baseflow conditions. The MCP simulations show smoother seasonal transitions, whereas the LSTM model sometimes produces more pronounced peak responses during the main melt period.

[95] In the *rainfall-dominated basins*, hydrographs are characterized by rapid runoff responses following precipitation events. Both MCP and LSTM models generally reproduce the timing of individual rainfall-driven peaks, demonstrating that the models successfully capture event-scale hydrological variability. Differences between the models appear mainly in peak magnitude and hydrograph shape. In some events the LSTM model produces sharper peaks, whereas in other cases the MCP simulations generate slightly higher peak flows with smoother recessions. The MCP hydrographs also tend to maintain more gradual recession limbs, reflecting the influence of storage and drainage processes within the mass-conserving framework.

[96] The comparison across dry, moderate, and wet years further demonstrates that both modeling approaches maintain consistent timing of major events across a wide range of hydrological conditions. Although deviations between simulated and observed flows increase during extreme wet-year events—when peak magnitudes become more difficult to reproduce — the overall hydrograph structure and event timing remain well represented by both models across the examined hydroclimatic regimes.

[97] Overall, the hydrograph analysis indicates that while the LSTM benchmark often produces sharper event responses, the MCP models capture the timing and temporal evolution of runoff events in a physically consistent manner. The smoother hydrograph behavior produced by the MCP framework reflects the explicit representation of storage dynamics, drainage processes, and mass conservation, which regulate the release of water during and after runoff events. These results demonstrate that the MCP models reproduce key features of event-scale hydrological behavior across diverse hydroclimatic regimes while maintaining physically-interpretable process representations.

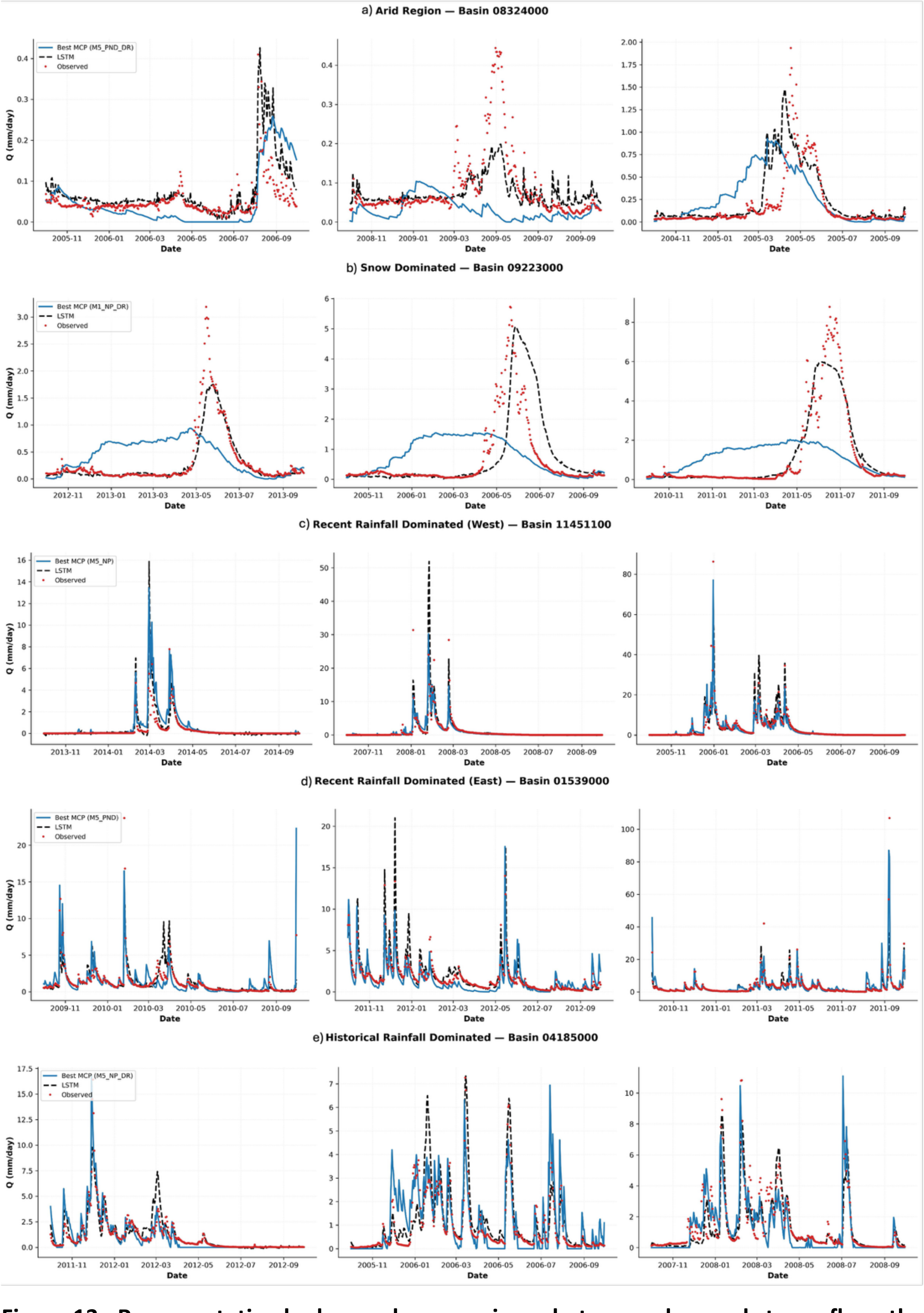

**Figure 13. Representative hydrograph comparisons between observed streamflow, the best-performing MCP configuration, and the LSTM benchmark for five basins representing major hydroclimatic regimes:** (a) arid region (USGS 08324000), (b) snow-dominated region (USGS 09223000), (c) recent rainfall-dominated region (West) (USGS 11451100), (d) recent rainfall-dominated region (East) (USGS 01539000), and (e) historical

rainfall-dominated region (USGS 04185000).For each basin, three hydrological years are shown to illustrate model behavior under contrasting hydrological conditions. **Left panels correspond to a relatively dry year, middle panels correspond to a year with moderate hydrological conditions, and right panels correspond to a wet year characterized by larger runoff events.** Observed streamflow is shown as red points, the best-performing MCP configuration for each basin is shown as a solid blue line, and the LSTM benchmark is shown as a dashed black line.

# 4 Discussion

## 4.1 Model Structural Complexity and Process Representation

[98] A key result of this study is that improvements in predictive skill arise primarily from embedding physically meaningful process constraints, rather than increasing model complexity alone. The largest performance gains occur when moving from the baseline configuration (M1) to early augmented models (M2–M3), indicating that a simple mass-conserving storage unit does not adequately represent dominant rainfall–runoff dynamics.

[99] Subsequent increases in structural complexity (M4–M5) generally yield smaller incremental improvements, suggesting *diminishing returns once key storage–release mechanisms are represented.* In some catchments, intermediate model structures perform comparably to, or better than, more complex variants, reinforcing the principle that *model structure should reflect dominant processes rather than maximize complexity.* Within the MCP framework, this progression is particularly informative because each structural addition corresponds to a *specific, interpretable process hypothesis*. As a result, performance differences across M1–M5 provide insight into which hydrological mechanisms most strongly control runoff behavior, rather than reflecting purely architectural changes.

[100] Overall, these findings suggest that hybrid physics–AI models benefit most from *targeted incorporation of dominant process representations*, especially those governing storage and release dynamics, while additional complexity plays a secondary role

## 4.2 Comparison Between Process-Based MCP and LSTM Models

[101] The comparison with the LSTM benchmark highlights a fundamental trade-off between *predictive flexibility and physical interpretability*. While LSTM models achieve more consistent performance across catchments, the MCP framework demonstrates that strong physical constraints do not preclude competitive predictive skill.

[102] The variability in MCP performance across basins suggests that its effectiveness depends on how well the embedded process representations align with dominant hydrological controls. In catchments where key processes are adequately represented, MCP models can match or exceed LSTM performance. In contrast, performance degrades when important mechanisms—such as snow storage and delayed melt—are not explicitly represented. This highlights a key distinction: *MCP performance is process-limited, whereas LSTM performance is data-limited but structurally unconstrained*.

[103] The Supporting Information (**Figures S3–S4**) further indicates that MCP structural augmentation primarily improves the dynamics of runoff response (e.g., correlation and variability), rather than systematically correcting bias. This suggests that *adding physical structure enhances how the model represents storage–release behavior* but does not automatically resolve all sources of model error.

[104] Beyond predictive performance, the two approaches differ fundamentally in interpretability. MCP internal states and fluxes correspond to physically meaningful quantities, enabling direct diagnosis of

model behavior in terms of hydrological processes. In contrast, LSTM representations remain latent, limiting the ability to attribute prediction errors to specific mechanisms.

[105] Taken together, these results suggest that MCP and LSTM models represent complementary modeling paradigms. LSTM models provide strong predictive robustness, while MCP models offer process transparency and mechanistic interpretability. The ability of MCP configurations to approach LSTM performance in many basins indicates that process-aware machine learning can bridge the gap between physically based and purely data-driven hydrological modeling.

### 4.3 Hydroclimatic and Event-Scale Controls on Model Performance

[106] Model performance is strongly controlled by hydroclimatic regime, indicating that the effectiveness of both structural augmentation and boundary-condition assumptions depends on dominant catchment processes. Rather than providing uniform improvements, the same model configuration can be beneficial in some regions and detrimental in others.

[107] A key result is *the contrasting role of vertical drainage across climates*. In *arid* and *snow-dominated basins*, enabling drainage consistently improves performance, suggesting that allowing water to leave the modeled storage is necessary to represent subsurface losses and delayed release pathways. In contrast, *rainfall-dominated basins* show the opposite behavior: drainage generally reduces performance, indicating that retaining water within the active storage better reflects runoff generation in humid environments. Compared to drainage, the effect of ponding is consistently minor, suggesting that lower-boundary conditions exert the dominant control on model behavior.

[108] These regime-dependent differences are also reflected in event-scale dynamics. Across basins, both MCP and LSTM models reproduce the timing of major runoff events, indicating that the dominant temporal variability of streamflow is well captured. However, systematic differences emerge in hydrograph shape. The LSTM tends to generate sharper peaks and more abrupt responses, whereas MCP simulations are generally smoother, with more gradual rises and recessions.

[109] This contrast highlights a fundamental difference in how the two models represent runoff generation. The smoother MCP hydrographs are consistent with a storage-controlled response, where water release is regulated by internal state and mass-conserving fluxes. In contrast, the sharper LSTM responses reflect its ability to reproduce rapid changes in discharge without explicit constraints on storage dynamics.

[110] The interaction between hydroclimate and event-scale behavior provides additional insight. In *arid systems*, MCP models capture the timing of episodic runoff events but tend to dampen peak magnitudes, consistent with their stronger storage control. In *snow-dominated basins*, the absence of an explicit snow storage leads to overly smooth seasonal transitions, limiting the model's ability to represent delayed melt-driven peaks. In *rainfall-dominated systems*, both models capture event timing well, but differences in recession behavior and low-flow persistence become more pronounced.

[111] Taken together, these results suggest that MCP performance is primarily limited by how well the embedded process representations capture storage persistence and release dynamics across wetness conditions. While the current formulation successfully reproduces many aspects of event-scale behavior, improving the representation of rapid runoff responses and prolonged low-flow persistence—particularly in *snow-influenced* and *humid systems*—remains an important direction for future development.

### 4.4 Process Interpretation Through Gate Response Functions

[112] An important advantage of the MCP framework is that the learned internal flux controls can be examined directly in physical terms. In particular, the subsurface gate response functions describe how

water release from the modeled storage unit varies as soil saturation increases. Instead of relying on a constant drainage coefficient or an uninterpretable latent state, the MCP framework learns a nonlinear mapping between internal saturation state and subsurface outflow activation. This provides a transparent representation of how the model regulates storage release across different wetness conditions.

[113] The learned gate-response curves show strong nonlinearity and substantial variation across model structures and boundary-condition scenarios (**Figure 14**). In most cases, subsurface gate activation is low under dry conditions and increases with increasing saturation, although the timing and steepness of this increase vary across configurations. Some curves remain near zero until relatively high saturation levels are reached and then increase rapidly, while others show a more gradual increase across a broader range of saturation values. This variation indicates that the MCP framework learns different state-dependent release behaviors across basins and structural configurations.

[114] *This behavior is hydrologically meaningful*. In classical soil physics, unsaturated and saturated conductivity are strongly nonlinear functions of pressure head or degree of saturation. Hydraulic conductivity typically remains small at low moisture content and increases rapidly as the soil approaches saturation. The learned MCP gate functions play an analogous role: they define an effective drainage response that varies with internal storage state. Although the MCP gate is not intended to reproduce a full Richards-equation constitutive relationship, it captures the same essential idea that water release should depend nonlinearly on wetness. In this sense, the gate can be interpreted as a learned effective conductivity function embedded within a mass-conserving neural architecture.

[115] **Figure 14** also shows that boundary-condition assumptions influence the learned gate shapes. In many basins, enabling vertical drainage shifts the response toward earlier or stronger activation across the saturation range. This indicates that the model adjusts its internal storage–release relationship when an additional lower-boundary outflow pathway is available. The ponding representation also alters the curvature of some responses, suggesting interactions between upper-boundary infiltration regulation and subsurface drainage behavior within the learned storage dynamics.

[116] Differences across model structures are equally informative. Lower-order models tend to exhibit simpler and more monotonic gate responses, whereas *higher-order models often show richer and more gradual transitions across saturation states*. This suggests that the progressive physical augmentation from M3 to M5 does not simply improve predictive skill numerically; it also *changes the character of the learned storage–release relationship*. As additional process realism is embedded in the unit, the gate response becomes capable of representing more nuanced transitions between dry, transitional, and wet states. This is consistent with the broader interpretation that higher-order MCP variants provide a more realistic internal hydrological response rather than merely increasing model flexibility.

[117] An important implication of these results is that the MCP framework offers a physically-interpretable alternative to dynamic parameter approaches commonly used in differentiable hydrology and hybrid ML models. In some data-driven frameworks, time-varying parameters are generated by neural networks but do not correspond to identifiable physical quantities. By contrast, the MCP gate remains tied to an explicit hydrological control variable—soil saturation—and represents a functional process relationship rather than an arbitrary latent adjustment. This makes it possible to interpret how the model is generating runoff, not just whether it predicts streamflow accurately.

[118] More broadly, the gate-response curves demonstrate that interpretability in hybrid hydrological models can extend beyond mass conservation or state labeling. The model can learn constitutive-style relationships that resemble familiar hydrological functions while remaining fully differentiable and trainable from data. This is one of the most promising aspects of the MCP framework: *it allows process hypotheses to be encoded, tested, and interpreted within a machine-learning setting*. The learned gate

functions therefore provide not only a useful diagnostic of model behavior, but also a pathway toward more physically grounded and scientifically informative AI-based hydrological models.

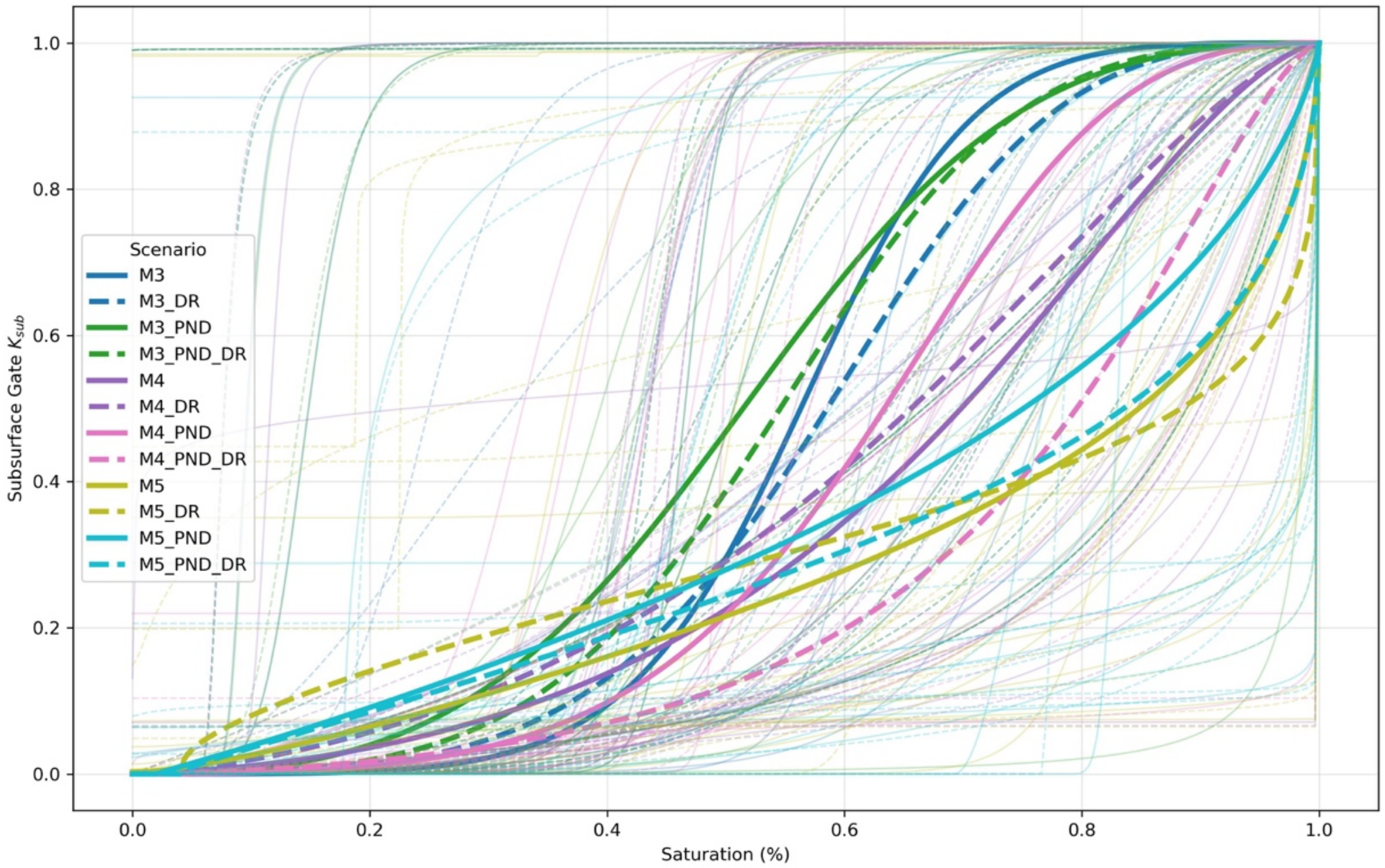

**Figure 14. Learned subsurface gate response functions showing the relationship between soil saturation and subsurface gate activation for MCP model configurations under different boundary-condition scenarios.** Thin lines represent individual catchment responses, while thick lines represent regional median behavior. Solid lines denote configurations without vertical drainage and dashed lines denote drainage-enabled configurations. Colors indicate model structures and ponding scenarios. The curves illustrate how the MCP framework learns nonlinear, state-dependent storage–release relationships that vary across model structures and boundary-condition assumptions.

### 4.5 Limitations and Future Work

[119] While the results demonstrate the potential of the MCP framework as an interpretable hybrid hydrological model, several limitations of the present study should be acknowledged.

[120] First, the experiments were conducted using a relatively small set of catchments spanning five hydroclimatic regimes. Although the selected basins provide diversity in climate and hydrological behavior, the total sample size remains limited compared with large-sample hydrological studies based on datasets such as CAMELS-US dataset. Evaluating the MCP framework across a larger number of basins would provide a more comprehensive assessment of its robustness and help determine whether the regional process patterns identified here hold across broader hydroclimatic gradients.

[121] Second, the current formulation focuses on a single effective soil storage unit and does not explicitly represent several hydrological processes that can influence catchment-scale runoff dynamics. In particular, *the framework does not include explicit representations of snow accumulation and melt processes, deeper groundwater storage, or channel routing*. These simplifications may contribute to some of the performance differences observed across hydroclimatic regimes, particularly in snow-dominated basins and in cases where low-flow persistence is difficult to reproduce. Incorporating additional storage

components or process representations could help address these limitations in future model developments.

[122] Third, the boundary-condition representations used in this study are intentionally simplified. Surface ponding is represented as a lumped temporary storage layer, and vertical drainage is implemented as a free lower-boundary flux leaving the modeled soil column. These simplified formulations are useful for isolating the influence of boundary conditions within the MCP framework, but they do not capture the full complexity of surface runoff routing or groundwater recharge processes. Future work could explore more physically detailed boundary representations, including lateral subsurface flow pathways or groundwater exchange dynamics.

[123] Another limitation concerns the spatial representation of hydrological processes. In the current implementation, the MCP framework operates at the catchment scale with a single storage unit representing an effective soil column. While this design simplifies interpretation and facilitates controlled testing of internal process hypotheses, real catchments exhibit substantial spatial heterogeneity in soils, topography, vegetation, and runoff-generation mechanisms. Extending the MCP architecture to include multiple interacting storage units or spatially distributed components may allow the framework to better represent this heterogeneity.

[124] Finally, the present study focuses on local models trained independently for each catchment. While this approach simplifies the interpretation of learned process relationships, it does not exploit potential information sharing across basins. Future research could investigate regional MCP formulations that incorporate catchment attributes or other forms of cross-basin learning to improve transferability while maintaining the physical interpretability of the internal storage–flux structure.

[125] Despite these limitations, the results demonstrate that embedding physically-interpretable process representations within a mass-conserving neural architecture is a promising direction for hybrid hydrological modeling. By combining machine-learning flexibility with explicit conservation constraints and interpretable internal dynamics, the MCP framework provides a pathway toward models that support both predictive performance and scientific insight into hydrological processes.

## 5 Conclusions

[126] This study investigated whether progressively embedding physically meaningful hydrological processes within a single *Mass-Conserving Perceptron* (MCP) storage unit can improve predictive performance and interpretability in rainfall–runoff modeling. By reformulating the MCP unit as a process-aware soil column and sequentially introducing bounded storage, state-dependent conductivity, infiltration controls, ponding storage, vertical drainage, and nonlinear water-table dynamics, we developed a hierarchy of physically-interpretable model structures. These models were evaluated across 15 catchments spanning five major hydroclimatic regimes of the continental United States and compared with a Long Short-Term Memory (LSTM) benchmark.

[127] The results show that progressively augmenting the internal physical structure of the MCP unit generally improves predictive performance. The largest improvements occur when key storage and drainage processes are introduced beyond the baseline configuration, while additional structural complexity yields smaller incremental gains. This pattern suggests that performance improvements arise primarily from incorporating the dominant hydrological mechanisms controlling storage–release behavior rather than from increasing model complexity alone.

[128] Boundary-condition assumptions were found to exert a strong and regionally dependent influence on model performance. Vertical drainage substantially improves predictive skill in arid and snow-dominated regions, indicating the importance of deep percolation and delayed subsurface storage

dynamics in these environments. In contrast, enabling free drainage generally reduces performance in rainfall-dominated catchments, where retaining water within the active storage better reproduces observed runoff generation. Surface ponding produces comparatively small effects across regions. These results demonstrate that the representation of boundary processes must be aligned with hydroclimatic context rather than applied uniformly across basins.

[129] Comparison with the LSTM benchmark shows that the best-performing MCP configurations approach the predictive skill of state-of-the-art deep learning models while maintaining explicit physical interpretability. Although the LSTM achieves slightly higher and more consistent performance across catchments, MCP models outperform the LSTM in several basins and reproduce many aspects of observed streamflow dynamics with comparable accuracy. Importantly, the MCP framework provides direct access to internal storage states and flux relationships, enabling mechanistic interpretation that is not possible with purely data-driven models.

[130] Analysis of the learned gate response functions further demonstrates that the MCP framework can learn physically meaningful relationships between storage state and subsurface water release. These nonlinear response curves resemble effective conductivity-type behaviors observed in soil hydrology and illustrate how interpretable process relationships can emerge within a differentiable machine-learning architecture. This capability highlights the potential of MCP-based models to serve not only as predictive tools but also as platforms for testing hydrological process hypotheses.

[131] Overall, the results suggest that hybrid physics–AI frameworks that embed physically-interpretable process structures within machine-learning architectures can achieve competitive predictive performance while retaining scientific transparency. Rather than treating machine learning and process-based modeling as competing paradigms, the MCP framework demonstrates a pathway for integrating the strengths of both approaches. Aligning model complexity with dominant hydrological processes and hydroclimatic context may therefore provide a promising strategy for developing next-generation hydrological models that support both accurate prediction and improved understanding of catchment dynamics. By embedding physically meaningful process representations directly within differentiable computational units, the MCP framework offers a promising foundation for interpretable and process-aware machine learning in hydrology.

[132] As always, we solicit and encourage constructive comments and debate on these and related aspects of geoscientific model development in the service of advancing scientific knowledge.

## Open Research

The data used in this study are freely available online:

## Acknowledgments

Funding for this project was provided by the National Oceanic and Atmospheric Administration (NOAA), awarded to the Cooperative Institute for Research on Hydrology (CIROH) through the NOAA Cooperative Agreement with The University of Alabama, NA22NWS4320003. This research was supported in part by an appointment to the Department of Defense (DOD) Research Participation Program administered by the Oak Ridge Institute for Science and Education (ORISE) through an interagency agreement between the U.S. Department of Energy (DOE) and the DOD. ORISE is managed by ORAU under DOE contract number DE-SC0014664. All opinions expressed in this paper are the author's and do not necessarily reflect the policies and views of DOD, DOE, or ORAU/ORISE.

**Table S1. Median predictive performance across the 15 study catchments measured using the $KGE_{ss}$ for all model structures and boundary-condition scenarios.**

| Model | $KGE_{ss}$ | Model | $KGE_{ss}$ | Model | $KGE_{ss}$ | Model | $KGE_{ss}$ |
|---|---|---|---|---|---|---|---|
| M1_NP | 0.62 | M1_PND | 0.62 | M1_NP_DR | 0.73 | M1_PND_DR | 0.73 |
| M2_NP | 0.71 | M2_PND | 0.71 | M2_NP_DR | 0.71 | M2_PND_DR | 0.72 |
| M3_NP | 0.71 | M3_PND | 0.71 | M3_NP_DR | 0.71 | M3_PND_DR | 0.71 |
| M4_NP | 0.74 | M4_PND | 0.74 | M4_NP_DR | 0.74 | M4_PND_DR | 0.73 |
| M5_NP | 0.77 | M5_PND | 0.76 | M5_NP_DR | 0.77 | M5_PND_DR | 0.76 |
| LSTM | 0.80 | | | | | | |

**Table S2.** Percentile statistics of predictive performance across the 15 study catchments measured using $KGE_{ss}$. The table summarizes the distribution of model performance for each configuration, including the minimum value, 5th percentile, quartiles, median, and 95th

percentile. The LSTM model exhibits higher lower-percentile performance, indicating more consistent skill across basins, while the upper percentiles of the MCP configurations approach the performance of the LSTM benchmark.

| Model | Worst KGE | $KGE_{SS}^{5\%}$ | $KGE_{SS}^{25\%}$ | $KGE_{SS}^{50\%}$ | $KGE_{SS}^{75\%}$ | $KGE_{SS}^{95\%}$ |
|---|---|---|---|---|---|---|
| M1_NP | 0.16 | 0.18 | 0.47 | 0.62 | 0.72 | 0.77 |
| M2_NP | 0.21 | 0.21 | 0.56 | 0.71 | 0.80 | 0.87 |
| M3_NP | 0.19 | 0.20 | 0.57 | 0.71 | 0.80 | 0.87 |
| M4_NP | 0.23 | 0.26 | 0.62 | 0.74 | 0.81 | 0.87 |
| M5_NP | 0.23 | 0.26 | 0.61 | 0.77 | 0.81 | 0.89 |
| M1_PND | 0.16 | 0.18 | 0.47 | 0.62 | 0.72 | 0.77 |
| M2_PND | 0.21 | 0.21 | 0.58 | 0.71 | 0.80 | 0.87 |
| M3_PND | 0.19 | 0.20 | 0.58 | 0.71 | 0.80 | 0.87 |
| M4_PND | 0.26 | 0.28 | 0.62 | 0.74 | 0.82 | 0.89 |
| M5_PND | 0.28 | 0.28 | 0.63 | 0.76 | 0.82 | 0.89 |
| M1_NP_DR | 0.20 | 0.20 | 0.49 | 0.73 | 0.75 | 0.82 |
| M2_NP_DR | 0.19 | 0.20 | 0.57 | 0.71 | 0.80 | 0.87 |
| M3_NP_DR | 0.19 | 0.20 | 0.59 | 0.71 | 0.80 | 0.87 |
| M4_NP_DR | 0.24 | 0.27 | 0.62 | 0.74 | 0.81 | 0.87 |
| M5_NP_DR | 0.24 | 0.27 | 0.63 | 0.77 | 0.81 | 0.89 |
| M1_PND_DR | 0.20 | 0.20 | 0.49 | 0.73 | 0.75 | 0.82 |
| M2_PND_DR | 0.19 | 0.20 | 0.58 | 0.72 | 0.80 | 0.87 |
| M3_PND_DR | 0.19 | 0.20 | 0.59 | 0.71 | 0.80 | 0.87 |
| M4_PND_DR | 0.27 | 0.28 | 0.62 | 0.73 | 0.82 | 0.88 |
| M5_PND_DR | 0.28 | 0.28 | 0.63 | 0.76 | 0.83 | 0.89 |
| LSTM | 0.69 | 0.72 | 0.76 | 0.80 | 0.85 | 0.92 |

**Table S3.** Incremental improvements in predictive performance relative to the baseline model (M1) measured as ΔKGE_{ss}. Statistics are computed across the 15 study catchments for models M2–M5 under four boundary-condition scenarios: standard without drainage (NP), ponding without drainage (PND), standard with drainage (NP_DR), and ponding with drainage (PND_DR).

| Scenario | Model | Median ΔKGE_{ss} | Mean ΔKGE_{ss} | Max ΔKGE_{ss} |
|---|---|---|---|---|
| NP | M2 | 0.08 | 0.08 | 0.28 |
| NP | M3 | 0.08 | 0.09 | 0.28 |
| NP | M4 | 0.09 | 0.11 | 0.26 |
| NP | M5 | 0.08 | 0.11 | 0.30 |
| PND | M2 | 0.08 | 0.09 | 0.28 |
| PND | M3 | 0.08 | 0.09 | 0.25 |
| PND | M4 | 0.10 | 0.11 | 0.29 |
| PND | M5 | 0.11 | 0.12 | 0.29 |
| NP_DR | M2 | 0.05 | 0.05 | 0.16 |
| NP_DR | M3 | 0.06 | 0.05 | 0.14 |
| NP_DR | M4 | 0.05 | 0.07 | 0.24 |
| NP_DR | M5 | 0.08 | 0.08 | 0.23 |
| PND_DR | M2 | 0.06 | 0.05 | 0.16 |
| PND_DR | M3 | 0.05 | 0.05 | 0.14 |
| PND_DR | M4 | 0.07 | 0.07 | 0.24 |
| PND_DR | M5 | 0.08 | 0.08 | 0.24 |

**Table S4.** Influence of boundary conditions on predictive performance for the best-performing model structure in each region. Differences are computed relative to the corresponding no-drainage or no-ponding configuration.

| Region | Best Model Structure | Best Median KGE | NP_DR − NP | PND_DR − PND | PND − NP | PND_DR − NP_DR |
|---|---|---|---|---|---|---|
| **Arid Region** | M4 | 0.79 | 0.06 | 0.07 | −0.008 | 0.01 |
| **Historical Rainfall Dominated** | M5 | 0.85 | −0.017 | −0.035 | 0.02 | −0.007 |
| **Recent Rainfall Dominated (East)** | M5 | 0.79 | −0.098 | −0.117 | 0.001 | 0.001 |
| **Recent Rainfall Dominated (West)** | M4 | 0.80 | −0.088 | −0.078 | 0.004 | 0.0004 |
| **Snow Dominated** | M5 | 0.71 | 0.41 | 0.42 | 0.01 | 0.01 |

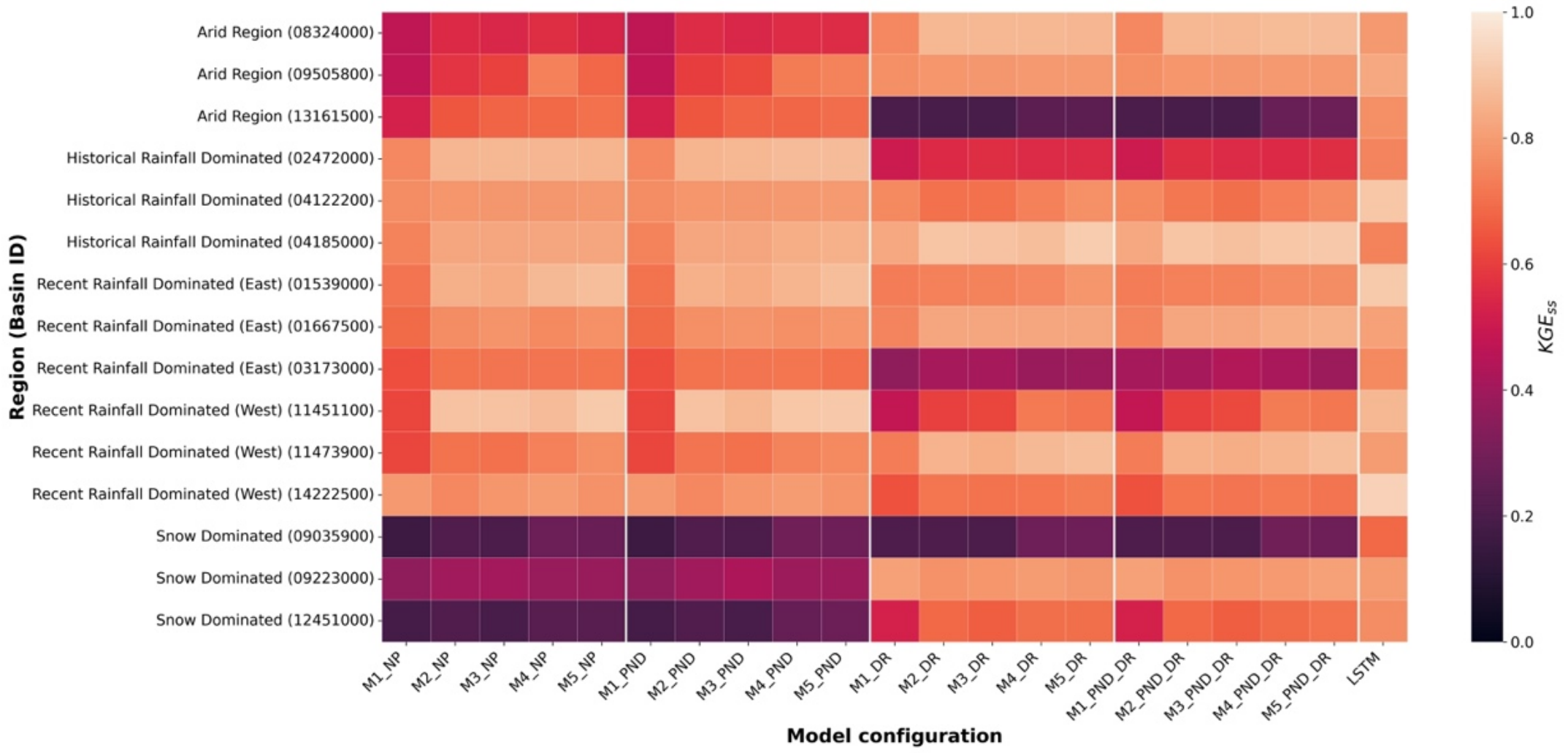

Figure S1. Basin-level comparison of predictive performance across model configurations. The heatmap shows $KGE_{ss}$ values for each model configuration across the 15 study catchments, grouped by hydroclimatic region. Columns represent the different model structures and boundary-condition scenarios (NP, PND, NP_DR, and PND_DR), while rows correspond to individual catchments. Warmer colors indicate higher predictive skill. The heatmap highlights systematic improvements in performance as model structural complexity increases from M1 to M5 and reveals substantial performance gains when vertical drainage processes are included.

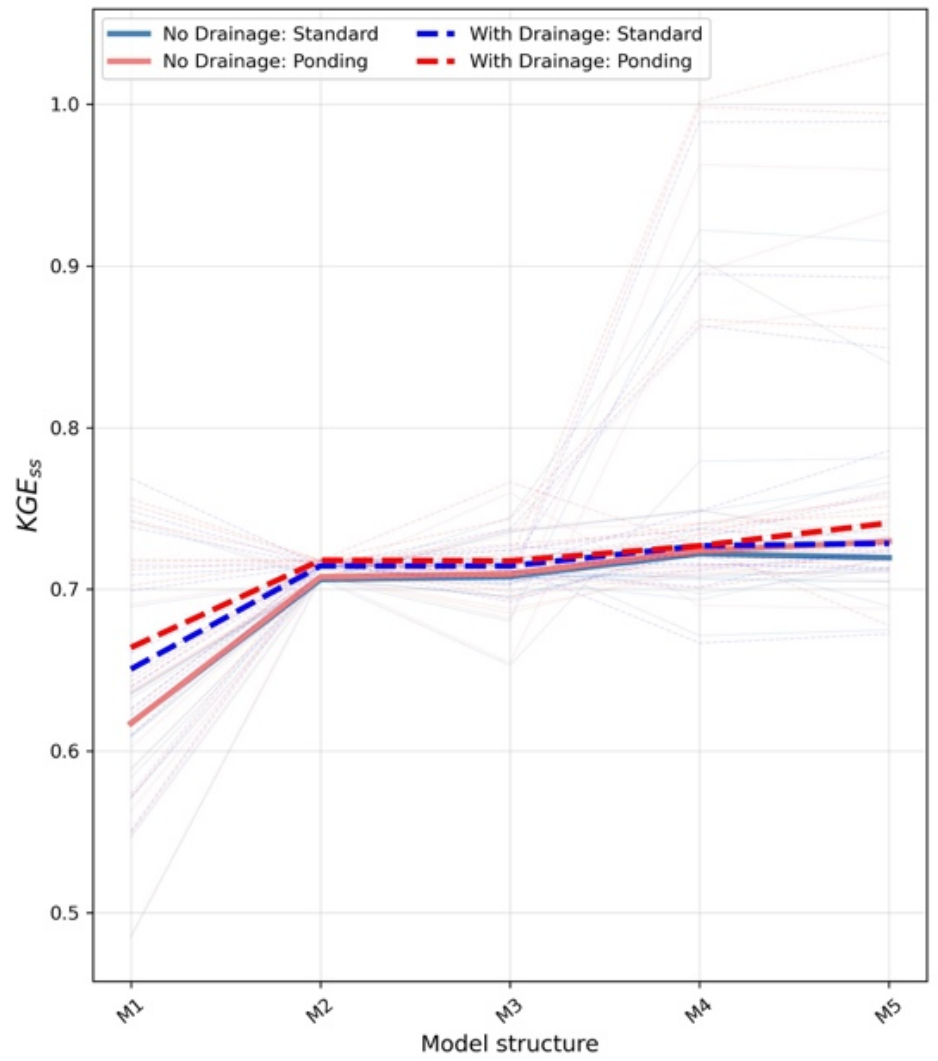

Figure S2. Evolution of predictive performance ($KGE_{ss}$) with increasing model structural complexity (M1–M5) across the 15 study catchments under four boundary-condition scenarios: NP, PND, NP_DR, and PND_DR. Thin lines represent individual basin responses, while bold lines indicate the median performance across all catchments. The figure illustrates how progressively augmenting the physical representation of hydrological processes improves predictive skill across basins.

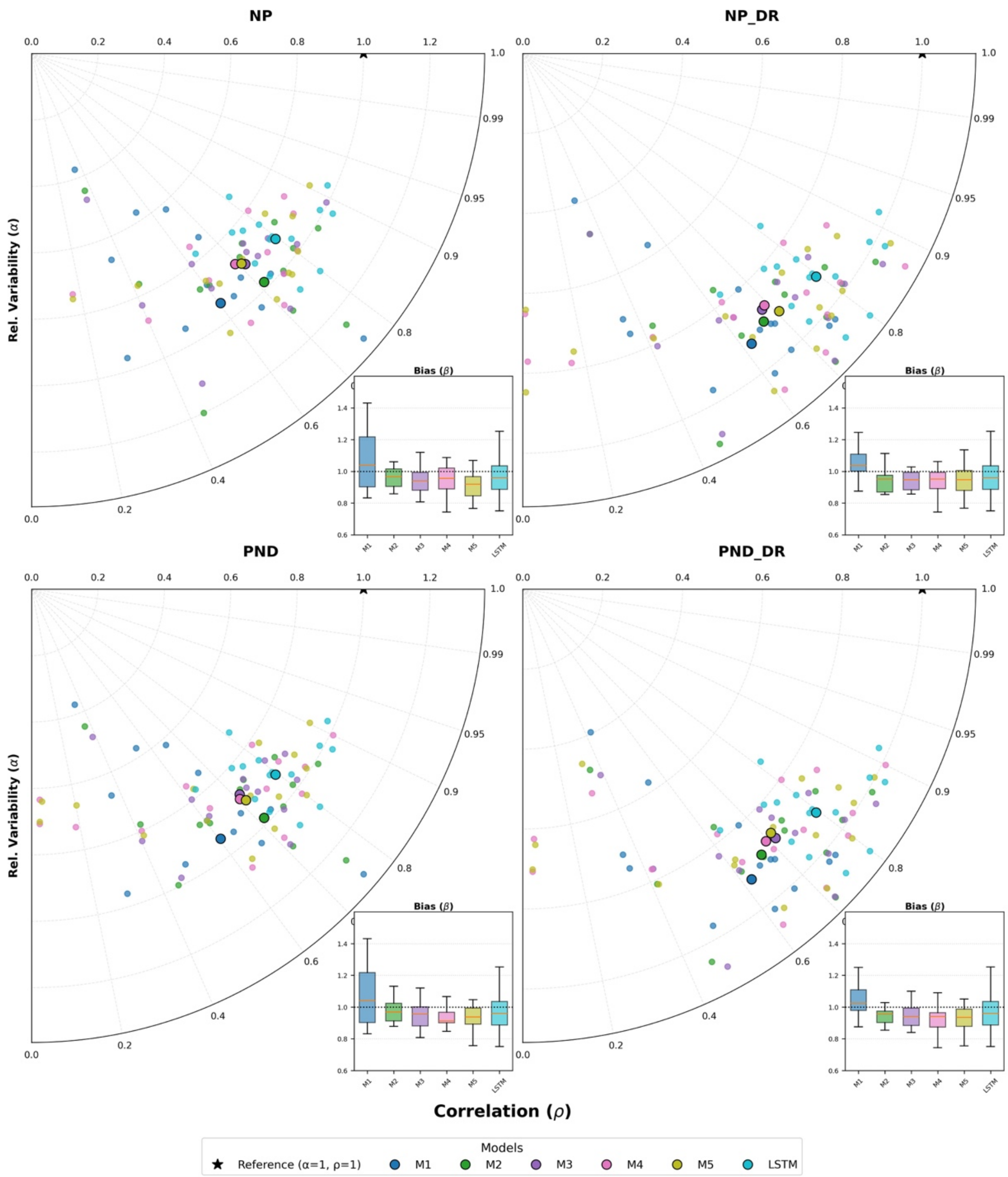

**Figure S3.** Taylor diagram comparing the performance of MCP model configurations and the LSTM benchmark across the study catchments. The diagram summarizes correlation, standard deviation, and root-mean-square error relative to observations.

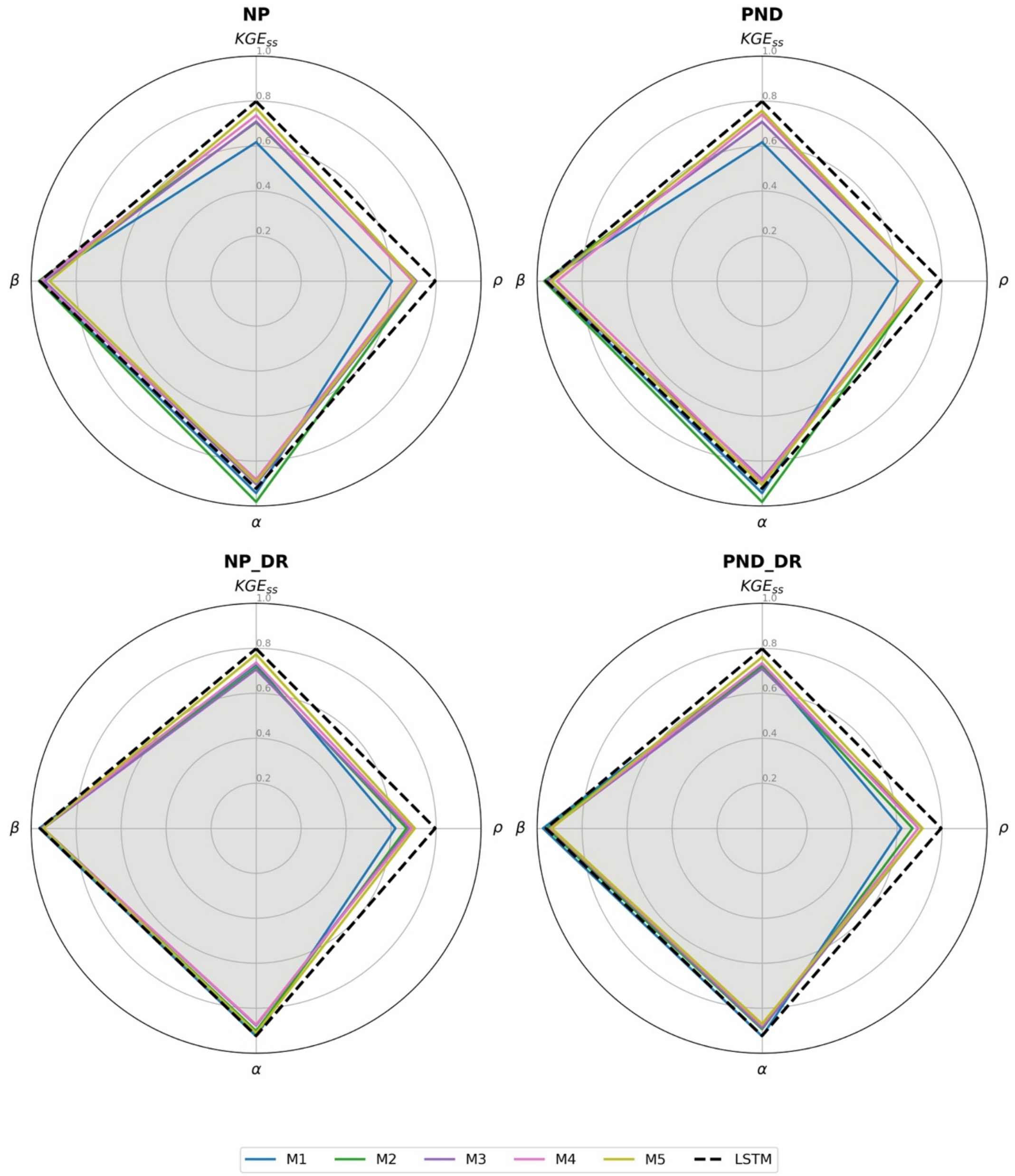

**Figure S4.** Radar plots showing the decomposition of the Kling–Gupta Efficiency components (correlation, variability ratio, and bias) for the MCP model configurations and the LSTM benchmark.